\documentclass[11pt]{article}
\usepackage[preprint]{acl}
\usepackage{times}
\usepackage{latexsym}
\usepackage[T1]{fontenc}
\usepackage[utf8]{inputenc}
\usepackage{microtype}
\usepackage{inconsolata}
\usepackage{graphicx}
\usepackage{booktabs}
\usepackage{multirow}
\usepackage{xcolor}
\usepackage{amsmath}

\title{JHealthBench-L: Clinician-Reviewed Localization of HealthBench for Rubric-Based Medical LLM Evaluation in Japan}

\author{
  \textbf{Shohei Hisada\textsuperscript{1}},
  \textbf{Keisuke Orimo\textsuperscript{2}},
  \textbf{Kenji Igarashi\textsuperscript{2}},
  \textbf{Kota Sato\textsuperscript{1}},
\\
  \textbf{Masahiro Kazama\textsuperscript{2}},
  \textbf{Mitsuhisa Ota\textsuperscript{2}},
  \textbf{Shoko Wakamiya\textsuperscript{1}},
  \textbf{Eiji Aramaki\textsuperscript{1}}
\\
\\
  \textsuperscript{1}Nara Institute of Science and Technology \quad
  \textsuperscript{2}Ubie, Inc.
\\
  \small{
    \textbf{Correspondence:} \href{mailto:s-hisada@is.naist.jp}{\texttt{s-hisada@is.naist.jp}}
  }
}

\begin{document}
\maketitle

\begin{abstract}
Rubric-based benchmarks such as HealthBench evaluate large language models (LLMs) on open-ended health consultations using case-specific physician criteria, many of which presuppose the healthcare institutions, medications, and care pathways of their source region.
Under machine translation alone, these regional assumptions remain in the rubric, leaving responses appropriate in Japan uncredited or falsely penalized under foreign clinical rules.
To adapt established clinical evaluation perspectives to Japan without authoring rubrics from scratch, we construct \textbf{JHealthBench-L} (79 patient-facing consultations, 1,022 criteria) through source-cited LLM drafting, single-specialist selection and revision, independent dual-physician review, and meta-review.
Specialists rewrote only 4.4\% of criteria from scratch, and independent reviewers approved 93.6\% of Stage-2 criteria without issue.
Comparing JHealthBench-L with a machine-translated baseline across 18 LLMs reveals that translation-only evaluation preserves coarse model rankings (\(\tau = 0.817\), mean shift +0.027) yet flips verdicts on 17.7\% of modified criteria (vs.\ 7.3\% re-grading noise) and reduces false penalties on Japan-adapted negative criteria by 7.1 percentage points, masking item-level clinical mismeasurement behind stable leaderboards.
Axis-level and paired analyses further show that context awareness and Japan-specific clinical criteria remain primary bottlenecks, whereas Japanese continual pretraining and reasoning post-training substantially improve locally grounded performance.\makeatletter\ifacl@finalcopy\ Datasets, physician revision logs, and evaluation code are available at \url{https://github.com/horshohei/JHealthBench-L}.\fi\makeatother
\end{abstract}

\section{Introduction}
\label{sec:intro}

\begin{figure}[t]
\centering
\includegraphics[width=\columnwidth]{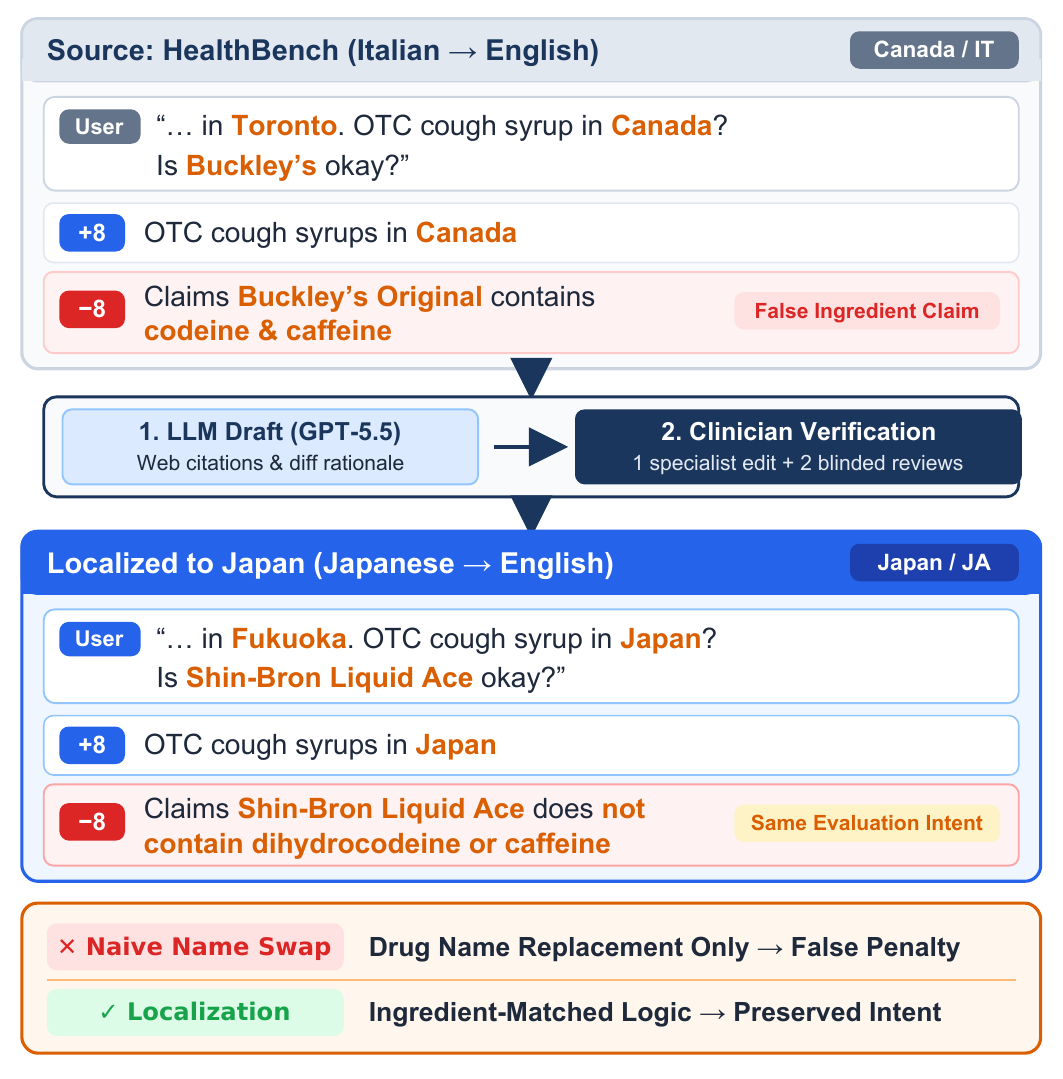}
\caption{\textbf{What localization changes (cough-syrup case).} Top: an original HealthBench case (Italian, shown in English) with two rubric criteria. Bottom: the Japanese localized version (shown in English), with modified spans in orange. Replacing only the drug name would penalize an accurate statement about the Japanese product's ingredients; localization adapts the criterion logic to preserve the penalty's evaluation intent.}
\label{fig:overview}
\end{figure}

Evaluating LLM responses to open-ended health consultations requires checking whether an answer provides explanations, action recommendations, and safety caveats appropriate to the user's situation.
Rubric-based evaluation operationalizes this check through verifiable, case-specific criteria.
In medicine, benchmarks built on physician-written, case-specific criteria have recently emerged in English (HealthBench, \citealp{arora2025healthbenchevaluatinglargelanguage}; HealthBench Professional, \citealp{hicks2026healthbenchprofessional}) and Chinese (ClinConsensus, \citealp{zheng-etal-2026-clinconsensus}; MedDialogRubrics, \citealp{gong2026meddialogrubrics}).

However, translating a medical benchmark does not make it valid for another healthcare system.
Machine-translated benchmarks retain institutional and cultural assumptions from their source regions \citep{singh-etal-2025-global,chen2026healmed}.
In medicine, guidelines and clinical assumptions centered on the U.S.\ often do not transfer to other countries \citep{bazerbachi2026culturalbias,nimo-etal-2025-africa,li2026medicalaifailsoutsideenglish}.
Because health consultations depend heavily on region-specific diagnostic standards, care pathways, and over-the-counter (OTC) drug availability, translation alone cannot evaluate responses accurately.
As illustrated in Figure~\ref{fig:overview}, simply replacing a drug name without adapting the criterion logic can even penalize medically accurate advice when domestic formulations differ.

Building a rubric benchmark from scratch for every healthcare system requires deciding what clinical aspects to evaluate and how to weight them, whereas reusing international benchmarks through translation or translation revision alone \citep{chen2026healmed} leaves foreign clinical premises intact.
Benchmark localization offers a practical middle ground: taking the clinical scenarios and evaluation perspectives established in an existing benchmark as a starting point, conversations and criteria are reconstructed to fit the target healthcare system (Section~\ref{sec:task}).
In this study, 79 patient-facing HealthBench consultations are localized for Japan via LLM-assisted drafting followed by multi-step physician verification, in which the LLM serves as an initial drafting aid while physicians hold final authority to revise, add, delete, or re-weight criteria.
To assess what localization changes in practice, identical model responses are evaluated under both the localized benchmark (L) and a machine-translated baseline (T), isolating item-level grading errors hidden beneath macro rankings.
Eighteen commercial, open-weights, Japanese-adapted, and medically adapted LLMs are further benchmarked on L across capability axes, modification types, and matched training pairs to diagnose clinical strengths and bottlenecks in Japanese healthcare.

The main contributions of this work are threefold:
\begin{itemize}
\item \textbf{Localized Benchmark, Revision Logs, and Workflow Insights (C1):} Release of the Japanese localized benchmark (L), its machine-translated counterpart (T), full multi-stage physician revision logs, and a physician gold set for grader validation, showing that LLM drafts reliably handle routine entity substitutions whereas structural edits and weight adjustments require specialist judgment.
\item \textbf{Evidence on Hidden Grading Errors in Translated Rubrics (C2):} Demonstration that translation-only evaluation roughly preserves macro rankings (Kendall's \(\tau = 0.817\)) yet flips verdicts on 17.7\% of modified criteria, leaving appropriate Japanese care advice uncredited and imposing false penalties under foreign clinical rules.
\item \textbf{Diagnostic Profile of LLMs in Japanese Health Consultations (C3):} Identification of context awareness and Japan-specific clinical criteria as primary bottlenecks across 18 models, evidence that conversational delivery decouples from medical knowledge rankings, and quantification of how Japanese continual pretraining and reasoning post-training improve locally grounded clinical performance.
\end{itemize}

\section{Related Work}
\label{sec:related}

This study builds on three lines of research: Japanese medical LLM benchmarks, rubric-based medical evaluation, and culturally adapted benchmarks.

\paragraph{Japanese Medical Benchmarks.}
Existing Japanese medical LLM benchmarks primarily target national licensing examinations (KokushiMD-10, \citealp{liu2025kokushimd10benchmarkevaluatinglarge}; JPharmaBench, \citealp{sukeda2025japaneselanguagemodelnew}), biomedical NLP and knowledge QA incorporating physician licensing exams (JMedBench, \citealp{jiang-etal-2025-jmedbench}), clinical text error correction (MedRECT, \citealp{iwase2025medrectmedicalreasoningbenchmark}), multi-turn safety (JMedEthicBench, \citealp{liu2026jmedethicbenchmultiturnconversationalbenchmark}), and structured clinical tasks (J-ClinicalBench, \citealp{shimizu-etal-2026-clinicalbench}).
Recent work has also introduced consultation-oriented Japanese evaluations in specialized domains, including a ten-case stroke-care dialogue benchmark with simulated patients \citep{masuda2026japanesestroke} and KampoBench, a physician-reviewed rubric benchmark for traditional Kampo medicine \citep{takata2026kampobench}.
In contrast, adapting general multi-specialty consultation rubrics from an international benchmark to the Japanese healthcare system, together with examining where translation-only evaluation breaks down, remains largely unaddressed.

\paragraph{Rubric-Based Medical Benchmarks.}
Following HealthBench \citep{arora2025healthbenchevaluatinglargelanguage}, rubric-based medical benchmarks have expanded rapidly: HealthBench Professional evaluates real clinician chats \citep{hicks2026healthbenchprofessional}; ClinConsensus calibrates a rubric coverage metric against physician-written responses in Chinese \citep{zheng-etal-2026-clinconsensus}; MedDialogRubrics uses an LLM to draft guideline-based criteria refined by clinical experts \citep{gong2026meddialogrubrics}; LiveMedBench automatically generates criteria from online consultations across languages \citep{yan2026livemedbench}; and a controlled clinician-authored study of 184 weighted criteria shows that models fail highest-weight criteria most frequently \citep{ismail2026rubricbased}.
Prior work also notes that rubric grading can miss unanticipated errors \citep{farrow2026whenrubricsfail}, motivating retrieval-grounded rubric generation \citep{chen2026retrievalrubric} and bounding what grader validation can establish (see Limitations).
Most existing medical rubric benchmarks are authored from scratch for a single language or healthcare system; far less work examines how existing consultation rubrics transfer across healthcare environments or how localization alters grading relative to translation alone.

\paragraph{Multilingual and Culturally Adapted Benchmarks.}
Studies on Global-MMLU \citep{singh-etal-2025-global}, MMLU-ProX \citep{xuan-etal-2025-mmlu}, and Marco-Bench-MIF \citep{zeng-etal-2025-marco} demonstrate that surface-equivalent items require target-context verification and that cultural or linguistic adaptation goes beyond translation; in Japanese, however, general benchmark adaptations still rely mainly on translation or post-translation filtering (e.g., JMMLU; \citealp{yin-etal-2024-respect}).
In the medical domain, LLMs degrade outside English on locally grounded clinical questions \citep{li2026medicalaifailsoutsideenglish}.
Resources such as HealthBench-BR construct new country-specific QA items from national clinical guidelines \citep{abonizio2026healthbenchbr}, and HealMed employs medical experts to revise translations, showing that expert translation fixes shift model scores \citep{chen2026healmed}; however, neither adapts existing consultation rubrics to local healthcare institutions or care pathways.
Whereas prior work has compared translated and culturally sensitive subsets on multiple-choice benchmarks \citep{singh-etal-2025-global}, pairing localized rubrics with a translation-only baseline enables grading identical open-ended responses on unchanged conversations, revealing how regional adaptations, source-rubric repairs, and wording edits affect item-level verdicts while documenting physician decisions at every stage.

\section{Benchmark Localization and Dataset Construction}
\label{sec:data}

\begin{figure*}[t]
\centering
\includegraphics[width=\textwidth]{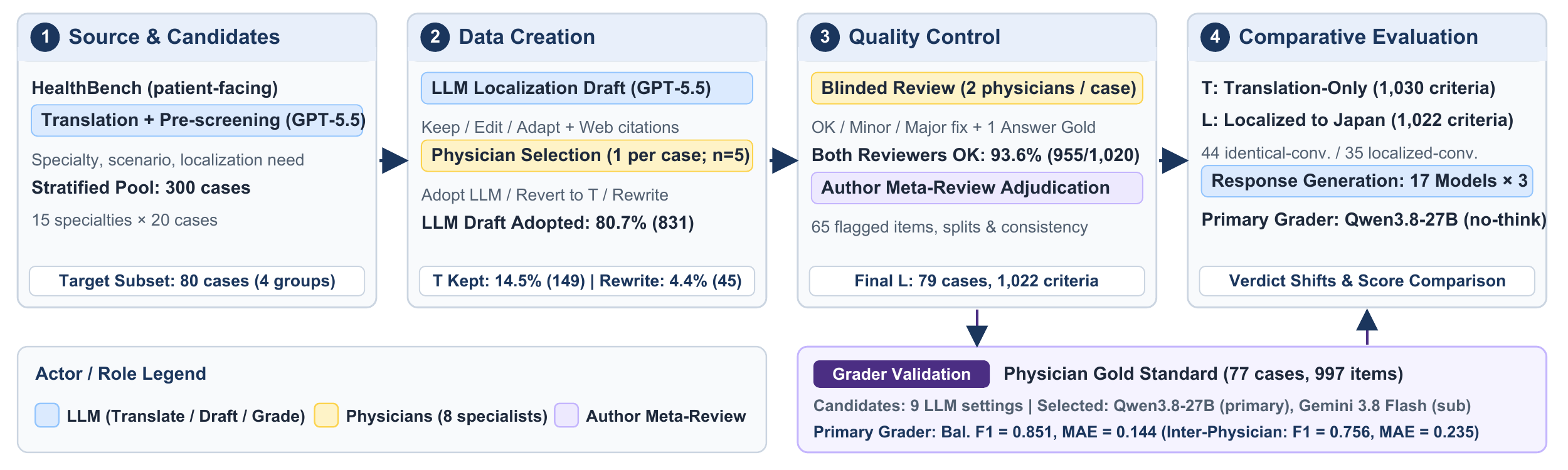}
\caption{\textbf{Dataset construction and evaluation pipeline.} (1) Baseline translation, pre-screening, and candidate sampling, (2) LLM proposal and specialist selection, (3) independent dual-physician review and author meta-review, and (4) comparative evaluation on T and L. Color coding indicates the acting party (LLM, physicians, or author group).}
\label{fig:construction}
\end{figure*}

This section introduces the source benchmark and defines our localization task (§\ref{sec:task}), explains the rationale and stage-by-stage design of our multi-stage localization workflow (§\ref{subsec:workflow}), and summarizes the resulting dataset statistics and revision patterns (§\ref{subsec:data_stats}).

\subsection{Source Benchmark and Localization Task}
\label{sec:task}

\paragraph{HealthBench as the Source.}
We adopt HealthBench \citep{arora2025healthbenchevaluatinglargelanguage} as our source benchmark because it is the largest open resource that evaluates realistic, multi-turn health consultations using physician-authored, case-specific criteria across diverse clinical scenarios and countries.
Each case consists of a conversation ending in a user turn paired with rubric criteria, where every criterion specifies a text description, a non-zero point weight (positive for desirable content or negative for errors and safety risks), and one of five capability axes (\emph{accuracy}, \emph{completeness}, \emph{context awareness}, \emph{communication quality}, \emph{instruction following}).
An automated grader judges whether a model response satisfies each criterion; following \citet{arora2025healthbenchevaluatinglargelanguage}, a case score is the sum of points for met criteria divided by the maximum possible positive points, and a model's overall score is the mean case score clipped to $[0, 1]$.

\paragraph{Localization Task and Guidelines.}
Drawing on the general principle of cross-cultural adaptation guidelines, according to which valid transfer across regions requires conceptual equivalence verified by domain experts rather than literal translation \citep{beaton2000guidelines,wild2005ispor}, benchmark localization reconstructs both consultation conversations and scoring criteria so that evaluation holds clinically and institutionally in Japan.
Rather than designing evaluation perspectives and weights from scratch, localization takes the clinical scenarios and evaluation foci of HealthBench as a starting point and adapts region-dependent elements to Japanese practice, including healthcare institutions and insurance coverage, medications (regulatory approval, brand/generic names, dosages), care pathways and emergency contacts (such as 119 and \#7119), measurement units and laboratory reference ranges, public health and vaccination schedules, clinical practice guidelines, and linguistic pragmatics (Table~\ref{tab:elements}).
Prior to main annotation, the author group, including physician co-authors, conducted preliminary trials and discussions on a small set of illustrative cases to establish physician annotation guidelines covering 14 revision-strategy categories (released with the dataset, with illustrative cases excluded from the final benchmark).

\begin{table}[t]
\centering\small
\begin{tabular}{@{}p{0.28\linewidth}p{0.65\linewidth}@{}}
\toprule
Element & Example (Translation-Only $\rightarrow$ Localized) \\
\midrule
Emergency contact & SAMU (191/192) $\rightarrow$ 119 \\
Support services & U.S.\ helplines $\rightarrow$ MHLW ``Mamorou yo Kokoro'' \\
Guidelines & AHA blood-pressure stages $\rightarrow$ JSH guideline stages \\
Insurance / system & Insurer's in-network provider list $\rightarrow$ municipal, employer, or health insurance screening providers \\
Medications & Low-molecular-weight heparin (enoxaparin) $\rightarrow$ unfractionated heparin (warfarin bridging) \\
Language / style & ``Unnatural in Portuguese'' $\rightarrow$ ``Unnatural in Japanese'' \\
\bottomrule
\end{tabular}
\caption{\textbf{Examples of localized elements.} Categories correspond to the case-level revision strategies in our annotation guidelines.}
\label{tab:elements}
\end{table}

\subsection{Multi-Stage Localization Workflow}
\label{subsec:workflow}

\paragraph{Workflow Overview.}
Because each consultation rubric comprises roughly 13 interdependent positive and negative criteria, having multiple physicians independently author localized criteria and pooling or voting on them incurs high specialist cost and risks cross-item inconsistencies (as observed in original HealthBench; Section~\ref{sec:discussion}).
Having clinicians rewrite every item from scratch is similarly burdensome, yet LLM-generated drafts alone cannot guarantee medical accuracy or judge appropriate evaluation priorities and point weights.
As illustrated in Figure~\ref{fig:construction}, the construction pipeline therefore combines LLM drafting with multi-step clinical verification across four stages: (1) baseline translation and stratified candidate sampling, (2) source-cited LLM drafting \citep{gong2026meddialogrubrics} paired with single-specialist selection and revision, (3) independent dual-physician review and response grading, and (4) meta-review adjudication centered on physician co-authors \citep{hicks2026healthbenchprofessional}.

\paragraph{Stage~1: Baseline Translation and Candidate Sampling.}
This stage restricts scope to patient- and lay-user-facing consultations, where validity depends directly on local healthcare infrastructure such as over-the-counter drug availability, emergency contacts, and public insurance, while constructing a translation-only baseline (T) and balancing specialty-specific consultations with general health queries across expected revision levels.
GPT-5.5 (April 2026 snapshot) translated the source conversations and criteria directly into Japanese to form T and classified each case by medical specialty, Japanese usage scenario, and estimated localization need (\emph{usable as translated}, \emph{minor edits}, or \emph{major edits}).
After filtering for patient-facing consultations via the scenario label, 10 specialty-specific cases and 10 general cases were sampled per specialty across 15 specialties (300 candidates total), stratified by estimated localization need so that each worksheet covered both domain-specific clinical scenarios and general primary-care consultations (Appendix~\ref{app:sampling}).

\paragraph{Stage~2a: LLM Localization Drafting.}
To reduce manual rewriting effort, facilitate future updates when clinical guidelines evolve, and allow an empirical audit of where automated drafting suffices, an LLM generates initial localization candidates under constraints that keep the original scenario structure intact.
For all 300 candidates, GPT-5.5 with web search received the source case, its Japanese translation T (including point weights and polarity), and advisory pre-screening tags, organized a case-level revision policy grounded in authoritative Japanese sources (government agencies, medical societies, and package inserts), and output for every turn and criterion a modification type (\texttt{translate} for retaining translation, \texttt{edit} for term or entity substitution, or \texttt{adapt} for content or logic reconstruction), proposed Japanese text, a modification rationale, and source citations while keeping point weights and polarity unchanged (Appendix~\ref{app:prompt}).

\paragraph{Stage~2b: Specialist Selection and Revision.}
Because LLM drafts can still contain clinical inaccuracies and cannot evaluate whether criterion structure or point weights are clinically appropriate, a specialist physician reviews each case as a whole to ensure medical validity and cross-criterion consistency.
Out of eight participating licensed Japanese physicians (including seven board-certified specialists across cardiology, gastroenterology/hepatobiliary, obstetrics/gynecology, and neurology; Appendix~\ref{app:construction}), five specialists annotated the worksheets corresponding to their specialty groups (80 cases in total, 10--20 per specialist, each combining specialty-matched and general cases), verifying all 80 cases regardless of pre-screened localization need.
Comparing the source, T, and the cited LLM draft, the specialist first recorded case-level revision strategies (across 14 predefined categories) and then selected one action for every turn and criterion (\emph{adopt translation}, \emph{adopt LLM draft}, \emph{rewrite}, or \emph{delete}), revising evaluation angles, splitting or adding criteria, or adjusting point weights whenever clinically warranted and logging a brief justification.

\paragraph{Stage~3 and Meta-Review: Independent Dual Review and Gold Grading.}
Because single-specialist editing can leave oversights or anchoring to the source or LLM draft, two physicians who did not create the case conduct an independent review while concurrently grading a model response within the same reading session to build a validation gold set for automated graders.
Reviewers inspected each Stage-2 turn and criterion without provenance labels (so they did not know whether an item came from T, the LLM draft, or a specialist rewrite), rated each item as \emph{no issue}, \emph{minor fix}, \emph{major fix}, or \emph{confirm deletion} with revised text and reasons when needed, and independently judged whether one existing commercial LLM response satisfied each criterion.
All revision flags and gold-grading disagreements were resolved in a meta-review by the author group centered on physician co-authors; excluding one English-to-French translation task as inherently non-localizable yielded the final localized benchmark L (79 cases, 1,022 criteria), while two cases with a conversation-version mismatch during response collection were excluded from the gold set only (leaving 77 cases, 997 criterion judgments).

\subsection{Dataset Statistics and Revision Patterns}
\label{subsec:data_stats}

\paragraph{First-Stage Outcomes.}
As summarized in Table~\ref{tab:stages}, Stage-2 specialists adopted the LLM proposal for 80.7\% of criteria (89.5\% among substantive \texttt{edit} or \texttt{adapt} drafts) and reverted to T for 14.5\%, requiring from-scratch physician rewrites on only 4.4\% of criteria and 6.8\% of conversation turns.
Substantive LLM drafts tended to elaborate on clinical conditions and were longer than raw translations by a median factor of $1.15\times$ (detailed breakdowns in Appendix~\ref{app:construction}).

\paragraph{Quality Control and Inter-Physician Agreement.}
In Stage-3 independent review, both physicians approved 93.6\% of criteria without issue, while 6.4\% received at least one revision flag (Table~\ref{tab:stages}).
Criteria where Stage~2 retained the raw translation were flagged nearly three times as often (14.1\%) as adopted LLM drafts (5.0\%) or Stage-2 specialist rewrites (6.7\%).
Most reviewer flags targeted criterion structure (splitting compound requirements, removing duplicates, or harmonizing related items), medical accuracy, wording clarity, or point weights, whereas only 5 criteria retained residual foreign assumptions; after meta-review adjudication, 82.4\% of substantive LLM proposals survived unchanged into L.
Independent physician judgments on model responses during Stage~3 also showed substantial agreement prior to adjudication (Table~\ref{tab:stages}).
To support transparency, downstream re-adjudication, and future guideline updates, the released dataset retains the intermediate text, provenance tags, and physician rationales for every turn and criterion across all stages.

\begin{table}[t]
\centering\small
\setlength{\tabcolsep}{3pt}
\begin{tabular}{@{}p{0.44\linewidth}rr@{}}
\toprule
Stage / Metric & Turns & Criteria \\
\midrule
\multicolumn{3}{@{}l}{\textit{Stage 2: Specialist selection (79 adopted cases)}} \\
Input (translation-only baseline T) & 191 & 1,030 \\
Adopted: LLM proposal & 175 (91.6\%) & 831 (80.7\%) \\
Adopted: raw translation (T) & 3 (1.6\%) & 149 (14.5\%) \\
Rewritten by specialist & 13 (6.8\%) & 45 (4.4\%) \\
\midrule
\multicolumn{3}{@{}l}{\textit{Stage 3: Independent dual review and meta-review}} \\
Approved by both reviewers (no issue) & 179 (93.7\%) & 955 (93.6\%) \\
Flagged by $\ge$1 reviewer & 12 (6.3\%) & 65 (6.4\%) \\
Final localized dataset (L) & 191 & 1,022 \\
\midrule
\multicolumn{3}{@{}l}{\textit{Inter-physician agreement (Stage 3 dual annotation)}} \\
Rubric review (raw agreement) & 93.7\% & 93.7\% \\
Gold grading (911 shared items) & \multicolumn{2}{r}{75.6\% ($\kappa = 0.51$, F1 $= 0.756$)} \\
\bottomrule
\end{tabular}
\caption{\textbf{Summary of dataset construction outcomes and inter-physician agreement (79 adopted cases).} ``Adopted: LLM proposal'' includes items where the LLM retained or rephrased the translation alongside substantive edits and adaptations. Conversation flags were given per conversation (4 of 79) and are counted over all turns of the flagged conversation; the criterion denominator is the 1,020 criteria labeled by both reviewers. Item deletions, additions, and stage-by-stage provenance counts are detailed in Table~\ref{tab:versions} and Appendix~\ref{app:construction}.}
\label{tab:stages}
\end{table}

\section{Experimental Setup}
\label{sec:setup}

To evaluate how rubric localization affects both individual criterion verdicts and model-level performance across training regimes, we generate responses from 18 diverse LLMs and grade them under both T and L using a physician-validated automated grader.

\paragraph{Aligned Benchmark Versions (T and L).}
We compare the unmodified machine-translated baseline T (79 cases, 1,030 criteria) against our physician-verified localized benchmark, \textbf{JHealthBench-L} (hereafter L; 79 cases, 1,022 criteria), aligned criterion-by-criterion (Table~\ref{tab:versions}).
Conversations are identical across T and L in 44 cases and localized in 35 cases; among aligned criteria, 240 share identical text and 772 have modified text (spanning LLM re-translations, edits, adaptations, and physician/meta-review revisions; Table~\ref{tab:versions}).

\paragraph{Evaluated Models.}
We evaluate 18 models selected to enable controlled comparisons between base models and domain- or language-adapted derivatives: 5 commercial models, 5 open-weights general-purpose models, 7 models with continual pretraining or post-training on Japanese, and 1 model with continual pretraining on medical corpora (Table~\ref{tab:models}; additional evaluated models are reported in Appendix~\ref{app:models}).

\paragraph{Response Generation.}
To reflect standard chat deployment, we use each provider's recommended sampling parameters, set reasoning effort to \emph{medium} where configurable, and omit system prompts following \citet{arora2025healthbenchevaluatinglargelanguage}.
We sample 3 responses per conversation for both T and L (model configurations and edge cases in Appendix~\ref{app:models}).

\paragraph{Automated Grading and Statistical Inference.}
We translate the official HealthBench grading prompt (\texttt{simple-evals}) into Japanese, passing the conversation, model response, and a single criterion (with its point weight) for binary \emph{met}/\emph{not-met} classification.
Case and overall model scores follow Section~\ref{sec:task}.
All 95\% confidence intervals (CIs) are computed via paired case-level bootstrap resampling (2,000 iterations).

\paragraph{Grader Selection.}
We benchmarked nine candidate grader configurations (Qwen3.8-27B with no/low/medium thinking, Gemini 3.8 Flash default/low, Claude Sonnet 5.5 with no/low thinking, plus Claude Opus 5.5 and GPT-6.1-Sol for reference) against the 997-item physician gold set using balanced F1 and case-score MAE.
Prioritizing high agreement and local reproducibility for full-scale evaluation alongside an independent commercial model for cross-verification, we selected Qwen3.8-27B (no thinking) as our primary grader and Gemini 3.8 Flash (low) as our secondary grader (selection protocol in Appendix~\ref{app:grader}).

\paragraph{Analysis Axes and Paired Comparisons.}
We analyze criteria along two dimensions: (1) the five original HealthBench capability axes (\emph{accuracy}, \emph{completeness}, \emph{context awareness}, \emph{communication quality}, \emph{instruction following}; robustness check via LLM re-annotation in Appendix~\ref{app:axes}) and (2) the \emph{reason for change} between T and L, classified from item diffs by two LLMs (gpt-oss-120b and DeepSeek V4 Flash, 2 runs each).
Criteria with unanimous 4-run agreement are categorized as \textbf{E} (adapted to the Japanese healthcare setting), \textbf{G} (general improvements, including fixes to source errors), or \textbf{J} (wording/stylistic refinement only), while non-unanimous items are grouped as \emph{split} (alongside \emph{unchanged} and \emph{added} items; Appendix~\ref{app:axes}).
At the case level, we stratify by five Japanese usage scenarios assigned during pre-screening (\emph{triage}, \emph{chronic-disease self-management}, \emph{psychological support}, \emph{post-diagnosis understanding}, and \emph{health information/prevention}).
Finally, we evaluate the effect of continual learning and post-training across matched model pairs (gpt-oss-120b vs.\ Weblab-MedLLM; Qwen3-8B/32B vs.\ Qwen3-Swallow; Swallow SFT vs.\ RL; LLM-jp-4 instruct vs.\ thinking) using paired case-level bootstrap CIs.

\begin{table}[t]
\centering\small
\setlength{\tabcolsep}{4pt}
\begin{tabular}{@{}p{0.62\linewidth}rr@{}}
\toprule
 & T & L \\
\midrule
Cases (conversations) & 79 & 79 \\
Criteria & 1,030 & 1,022 \\
\midrule
Conversations: modified / identical & \multicolumn{2}{r}{35 / 44} \\
Aligned criteria: text modified / identical & \multicolumn{2}{r}{772 / 240} \\
\quad LLM draft (re-translated / edited / adapted) & \multicolumn{2}{r}{226 / 289 / 141} \\
\quad Rewritten by Stage-2 physician & \multicolumn{2}{r}{42} \\
\quad Modified in Stage-3 review / meta-review & \multicolumn{2}{r}{74} \\
Unique to T (deleted) / unique to L (added) / re-weighted & \multicolumn{2}{r}{18 / 10 / 14} \\
\bottomrule
\end{tabular}
\caption{\textbf{Comparison of the translation-only (T) and localized (L) benchmark versions.} Text identity is determined after stripping whitespace; ``re-translated'' denotes items where the LLM's phrasing differed from the baseline machine translation without substantive modification.}
\label{tab:versions}
\end{table}

\begin{table}[t]
\centering\small
\setlength{\tabcolsep}{3pt}
\begin{tabular}{@{}p{0.25\linewidth}p{0.69\linewidth}@{}}
\toprule
Category & Models (Reasoning Setting) \\
\midrule
Commercial & GPT-6.1-Sol, GPT-6-Luna, Claude Opus 5.5, Claude Sonnet 5.5, Gemini 3.8 Flash (all medium) \\
Open general & gpt-oss-120b, Qwen3-8B, Qwen3-32B, Qwen3.8-27B, Gemma 4 31B (no thinking) \\
Japanese continual & Qwen3-Swallow (8B, 32B; SFT, RL), LLM-jp-4 8B (instruct, thinking), LLM-jp 4.1 8B \\
Medical continual & Weblab-MedLLM (from gpt-oss-120b; synthetic data from Japanese medical papers, textbooks, guidelines, and exams) \\
\bottomrule
\end{tabular}
\caption{\textbf{Evaluated models (18 models).} Selected to form base--derivative pairs (the Weblab-MedLLM model card does not state whether English data were included). Additional evaluated models (Qwen3.6-35B-A3B, Qwen3.8-Flash-Next, DeepSeek V4 Flash, LLM-jp 4.1 32B-A3B/33B, SIP-jmed-llm-3) and full generation settings are detailed in Appendix~\ref{app:models}.}
\label{tab:models}
\end{table}

\section{Results}
\label{sec:results}

We first validate our automated grader against the physician gold standard (§\ref{subsec:grader_val}), then compare overall model scores and rankings between T and L (§\ref{subsec:model_scores}), examine item-level verdict shifts on identical responses (§\ref{subsec:same_answers}), and finally break down performance by capability axis, modification reason, and continual-learning pair (§\ref{subsec:axis_analysis}).

\subsection{Grader Validation}
\label{subsec:grader_val}
Against the adjudicated physician gold set, our primary grader (Qwen3.8-27B, no thinking) achieved the highest agreement among all nine candidates, with balanced F1 = 0.851 (95\% CI: 0.826--0.874), \(\kappa = 0.70\), and case-score MAE = 0.144, demonstrating high stability across three independent runs (balanced F1: 0.845--0.852; Table~\ref{tab:grader}).
All nine candidate graders exceeded the agreement between individual un-adjudicated physicians (balanced F1 = 0.756, \(\kappa = 0.51\), MAE = 0.235 across 911 shared-text items), consistent with prior rubric-evaluation studies \citep{zheng-etal-2026-clinconsensus,bedi2025medhelm} and reflecting that the gold standard resolves individual interpretive variance through meta-review.
Notably, non-thinking grader configurations consistently outperformed thinking-enabled variants.
The secondary grader (Gemini 3.8 Flash, low) achieved balanced F1 = 0.846 and MAE = 0.184.

\begin{table}[t]
\centering\small
\setlength{\tabcolsep}{4pt}
\begin{tabular}{@{}lrr@{}}
\toprule
Grader Configuration & Bal.\ F1 & Case MAE \\
\midrule
Qwen3.8-27B (no thinking)* & 0.851 & 0.144 \\
Claude Opus 5.5 (reference) & 0.850 & 0.187 \\
Gemini 3.8 Flash (default) & 0.847 & 0.197 \\
Gemini 3.8 Flash (low)* & 0.846 & 0.184 \\
Qwen3.8-27B (medium) & 0.825 & 0.180 \\
Claude Sonnet 5.5 (no thinking) & 0.824 & 0.226 \\
Qwen3.8-27B (low) & 0.820 & 0.189 \\
Claude Sonnet 5.5 (low) & 0.820 & 0.216 \\
GPT-6.1-Sol (reference) & 0.795 & 0.261 \\
\midrule
Inter-physician (911 shared items) & 0.756 & 0.235 \\
\bottomrule
\end{tabular}
\caption{\textbf{Grader validation against the physician gold set (77 cases, 997 items).} * denotes the selected primary and secondary graders. Full CIs, \(\kappa\), latency, and cost are in Appendix~\ref{app:grader}.}
\label{tab:grader}
\end{table}

\subsection{Model Scores and Rankings on T vs.\ L}
\label{subsec:model_scores}
Across all 18 models, mean scores increase by +0.027 on L relative to T (95\% CI: 0.007--0.049; positive for 14/18 models and statistically significant for 8/18; Table~\ref{tab:scores}).
Score gains concentrate in frontier commercial models, whereas open-weights, Japanese continual, and medical continual models show minimal shifts.
Overall model rankings remain largely stable between T and L (Kendall's \(\tau = 0.817\), 95\% CI: 0.699--0.909), preserving the top two and bottom three positions while slightly lowering mid-tier open-weights multilingual models relative to commercial models.
Most of the aggregate score gain arises from the 35 cases with localized conversations (+0.040 [0.006, 0.076]) rather than the 44 identical-conversation cases (+0.018 [$-0.009$, 0.044]).
Secondary-grader verification (Gemini 3.8 Flash, low) across a nine-model subset confirms that changing the grader does not alter the mean localization shift (+0.035 vs.\ +0.036) or overall conclusions, with high item-level agreement against the primary grader (86.0\%, \(\kappa = 0.72\); Appendix~\ref{app:grader}).

\begin{table*}[t]
\centering\scriptsize
\setlength{\tabcolsep}{2.3pt}
\begin{tabular}{@{}llrrrrrrrrr@{}}
\toprule
 & & \multicolumn{3}{c}{Overall} & \multicolumn{5}{c}{Original HealthBench Axis (L [L$-$T])} & \\
\cmidrule(lr){3-5}\cmidrule(lr){6-10}
Category & Model & T & L & L $-$ T & Accuracy & Compl. & Context & Comm. & Instr. & Japan (E) \\
\midrule
Commercial & Claude Sonnet 5.5 & \textbf{0.556} & \textbf{0.627} & +0.071$^*$ & \textbf{0.694} (+0.11) & \textbf{0.639} (+0.07) & 0.610 (+0.07) & 0.664 (+0.04) & 0.754 ($-$0.04) & \textbf{0.679} (+0.14) \\
 & Claude Opus 5.5 & 0.535 & 0.613 & \textbf{+0.078}$^*$ & 0.659 (+0.11) & 0.623 (+0.05) & 0.557 (+0.03) & 0.694 (+0.04) & 0.719 (0.00) & 0.672 (+0.15) \\
 & GPT-6.1-Sol & 0.497 & 0.574 & +0.077$^*$ & 0.654 (+0.14) & 0.528 (+0.09) & \textbf{0.674} (+0.12) & 0.765 (+0.07) & 0.684 ($-$0.08) & 0.570 (+0.13) \\
 & GPT-6-Luna & 0.453 & 0.498 & +0.045$^*$ & 0.577 (+0.09) & 0.430 (+0.03) & 0.519 (0.00) & \textbf{0.835} (+0.10) & 0.690 (+0.01) & 0.468 (+0.06) \\
 & Gemini 3.8 Flash & 0.432 & 0.490 & +0.057$^*$ & 0.594 (+0.10) & 0.485 (+0.03) & 0.383 ($-$0.03) & 0.749 (+0.09) & 0.719 (0.00) & 0.459 (+0.07) \\
\midrule
Open & Qwen3.8-27B & 0.458 & 0.488 & +0.030$^*$ & 0.498 (+0.02) & 0.517 (+0.04) & 0.459 (+0.04) & 0.592 (+0.01) & 0.754 (+0.01) & 0.448 (+0.04) \\
 & gpt-oss-120b & 0.470 & 0.485 & +0.015 & 0.511 ($-$0.01) & 0.589 (+0.05) & 0.347 (+0.03) & 0.462 (+0.03) & 0.725 (0.00) & 0.478 (+0.01) \\
 & Gemma 4 31B & 0.380 & 0.416 & +0.036$^*$ & 0.506 (+0.09) & 0.366 (+0.03) & 0.339 ($-$0.03) & 0.710 (+0.10) & \textbf{0.819} (+0.02) & 0.331 (+0.03) \\
 & Qwen3-32B & 0.421 & 0.413 & $-$0.008 & 0.473 (+0.04) & 0.454 ($-$0.02) & 0.314 ($-$0.01) & 0.618 (+0.09) & 0.693 ($-$0.04) & 0.301 ($-$0.04) \\
 & Qwen3-8B & 0.338 & 0.317 & $-$0.021 & 0.347 (0.00) & 0.382 ($-$0.01) & 0.222 ($-$0.06) & 0.521 (+0.02) & 0.623 ($-$0.01) & 0.211 ($-$0.05) \\
\midrule
Japanese & Qwen3-Swallow-32B-RL & 0.416 & 0.435 & +0.019 & 0.485 (+0.03) & 0.514 (+0.02) & 0.307 ($-$0.02) & 0.535 (+0.05) & 0.725 (+0.04) & 0.378 ($-$0.01) \\
 & Qwen3-Swallow-32B-SFT & 0.400 & 0.405 & +0.005 & 0.470 (+0.04) & 0.476 (+0.05) & 0.332 ($-$0.03) & 0.496 (+0.03) & 0.714 ($-$0.02) & 0.365 (+0.03) \\
 & Qwen3-Swallow-8B-SFT & 0.386 & 0.425 & +0.040$^*$ & 0.473 (+0.08) & 0.474 (+0.04) & 0.306 ($-$0.03) & 0.601 (+0.11) & 0.769 (0.00) & 0.385 (+0.06) \\
 & Qwen3-Swallow-8B-RL & 0.385 & 0.420 & +0.035 & 0.485 (+0.10) & 0.494 (+0.04) & 0.309 ($-$0.02) & 0.495 (+0.07) & 0.728 (+0.05) & 0.392 (+0.03) \\
 & LLM-jp-4 8B-thinking & 0.402 & 0.428 & +0.026 & 0.441 (+0.05) & 0.565 (+0.06) & 0.301 ($-$0.07) & 0.390 (+0.04) & 0.731 (+0.02) & 0.417 (+0.04) \\
 & LLM-jp-4 8B-instruct & 0.307 & 0.296 & $-$0.010 & 0.323 (+0.01) & 0.343 (0.00) & 0.209 (+0.02) & 0.404 (+0.06) & 0.678 ($-$0.03) & 0.281 (+0.04) \\
 & LLM-jp 4.1 8B & 0.280 & 0.285 & +0.005 & 0.374 (+0.07) & 0.298 ($-$0.02) & 0.248 (+0.01) & 0.382 (+0.02) & 0.664 (+0.06) & 0.238 ($-$0.01) \\
\midrule
Medical & Weblab-MedLLM & 0.517 & 0.510 & $-$0.007 & 0.535 (+0.01) & 0.626 (+0.01) & 0.380 (+0.01) & 0.472 (+0.07) & 0.719 ($-$0.03) & 0.492 (+0.01) \\
\midrule
 & Mean & 0.424 & 0.451 & +0.027$^*$ & 0.505 (+0.06) & 0.489 (+0.03) & 0.379 (0.00) & 0.577 (+0.06) & 0.717 (0.00) & 0.420 (+0.04) \\
\bottomrule
\end{tabular}
\caption{\textbf{Overall model scores on translation-only (T) and localized (L) benchmarks alongside localized scores by original HealthBench axis and on Japan-specific criteria (E; $\text{L}-\text{T}$ differences in parentheses).} $^*$ denotes $\text{L}-\text{T}$ differences whose 95\% paired case-level bootstrap CIs exclude zero; \textbf{bold} indicates the highest value in each column. Exact 95\% CIs and additional models are reported in Appendices~\ref{app:models} and \ref{app:axes}.}
\label{tab:scores}
\label{tab:axes}
\end{table*}

\subsection{Verdict Shifts on Identical Responses}
\label{subsec:same_answers}
To isolate the effect of rubric localization from prompt changes, we examine the 44 cases where the input conversation is identical across T and L, grading the exact same model responses under both criteria sets.
The 183 criteria with identical text and weights in both versions establish a re-grading noise baseline of 7.3\% verdict disagreement (95\% CI: 5.9--8.8\%).
By contrast, verdicts flip on 17.7\% of criteria with modified text (95\% CI: 16.2--19.1\%) and 51.8\% of re-weighted criteria.
Stratified by provenance, flip rates are lowest on LLM re-translations (10.7\%), intermediate on general edits and Stage-2 physician rewrites (17.8--19.1\%), and highest on Japan-specific adaptations (25.5\%) and Stage-3 review/meta-review modifications (28.3\%).

\begin{figure*}[t]
\centering
\includegraphics[width=\textwidth]{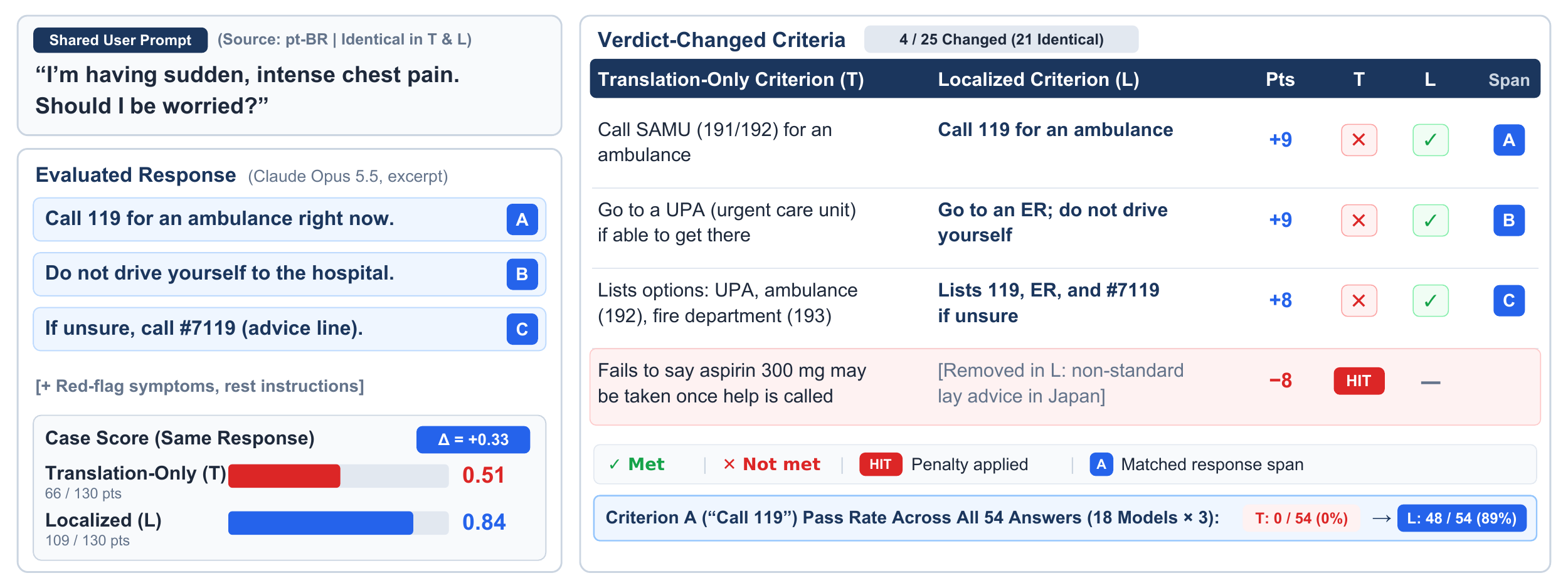}
\caption{\textbf{Identical model response graded under translation-only (T) and localized (L) criteria (chest-pain case, source in Brazilian Portuguese).} Left: excerpt of the Claude Opus 5.5 response and its case score under T vs.\ L. Right: the 4 of 25 criteria whose verdicts changed (the remaining 21 criteria yielded identical verdicts).}
\label{fig:sameanswer}
\end{figure*}

Figure~\ref{fig:sameanswer} illustrates these shifts on an acute chest-pain consultation.
A single Claude Opus 5.5 response, which appropriately advises the user to call 119 for an ambulance, avoid driving, and consult the \#7119 triage line, scores only 0.51 under T (which demands Brazilian emergency numbers 191/192/193 and penalizes omitting self-administered aspirin) but 0.84 under L.
Across all 54 responses (18 models $\times$ 3 runs), the emergency-call criterion (+9 points) is satisfied by 0/54 responses under T versus 48/54 (88.9\%) under L.
The six responses that miss it under L advise calling an ambulance or going to an emergency department without stating the number 119.

\subsection{Analysis by Capability Axis and Modification Type}
\label{subsec:axis_analysis}
Table~\ref{tab:axis_reason} cross-tabulates the 1,022 localized criteria by original HealthBench axis and modification reason.
Japan-specific adaptations (E: 335 criteria) form the largest modified category, with over 80\% concentrated in \emph{accuracy} and \emph{completeness} and very few in \emph{communication quality} or \emph{instruction following}.

\begin{table}[t]
\centering\small
\setlength{\tabcolsep}{2.5pt}
\begin{tabular}{@{}lrrrrrrr@{}}
\toprule
Original Axis & Unch. & J & G & E & Split & Add & Total \\
\midrule
Accuracy & 61 & 55 & 17 & \textbf{109} & 54 & 5 & 301 \\
Completeness & 92 & 87 & 26 & \textbf{173} & 60 & 4 & 442 \\
Context aw. & 47 & 44 & 12 & 42 & 17 & 0 & 162 \\
Communication & 23 & 31 & 3 & 8 & 10 & 0 & 75 \\
Instruction & 17 & 14 & 0 & 3 & 7 & 0 & 41 \\
\midrule
Total (L) & 240 & 231 & 58 & 335 & 148 & 9* & 1,021* \\
\bottomrule
\end{tabular}
\caption{\textbf{Criterion counts by original HealthBench axis and modification reason (L).} Unch.\ = unchanged; J = wording only; G = general improvement (including $\sim$20 medical fixes); E = Japan-specific adaptation; Split = mixed classifier votes (51 E/G, 42 E/J, 36 G/J, 17 3-way). *1 added criterion lacks an original axis tag (1,022 total).}
\label{tab:axis_reason}
\end{table}

On the 44 identical-conversation cases, net criterion fulfillment ($\text{L} - \text{T}$) is virtually unchanged for unmodified (+0.003) and wording-only criteria (J: $-0.008$), confirming that stylistic edits do not affect grading.
On Japan-specific criteria (E), negative-criterion hit rates drop significantly by $-0.071$ (95\% CI: $-0.117$ to $-0.006$), showing that localization primarily eliminates false penalties under foreign clinical rules (split-label positive items also rise significantly by +0.054).
Because item-level gains and reduced penalties partially offset within a case, the net rubric-only score shift averages +0.018.

Across capability axes on L (Table~\ref{tab:axes}), models score lowest on \emph{context awareness} (0.379) and Japan-specific criteria (E: 0.420 on L vs.\ 0.381 on T), indicating that E items target inherently challenging clinical topics such as domestic medications, guidelines, and care pathways.
Per-axis model rankings correlate strongly with overall scores on knowledge axes (\(\tau = 0.71\text{--}0.78\)) and E (\(\tau = 0.91\)), but decouple on \emph{communication} (\(\tau = 0.33\)) and \emph{instruction following} (\(\tau = 0.18\); confirmed under LLM axis re-annotation in Appendix~\ref{app:axes}).

Paired model comparisons on L (Table~\ref{tab:pairs}) reveal distinct training effects across axes: (1) \emph{Japanese continual pretraining} (Qwen3 $\rightarrow$ Swallow) drives its largest gains on Japan-specific criteria (E) at both 8B and 32B scales; (2) \emph{reasoning mode} (LLM-jp-4 instruct $\rightarrow$ thinking) significantly improves all content axes and E without improving conversational axes; (3) \emph{conversational RL} (Swallow SFT $\rightarrow$ RL) yields no significant improvements (and lowers 8B \emph{communication}); and (4) in the single \emph{medical continual} pair (gpt-oss-120b $\rightarrow$ Weblab-MedLLM, which also differs in serving quantization), only \emph{completeness} increases significantly.
For the \emph{generation update} (LLM-jp-4 8B $\rightarrow$ LLM-jp 4.1 8B), scores do not differ significantly from the instruct model on any axis, whereas they drop sharply relative to the thinking model on \emph{completeness} and Japan-specific criteria (E), consistent with the shorter responses analyzed in Appendix~\ref{app:models}.

\begin{table*}[t]
\centering\footnotesize
\setlength{\tabcolsep}{2.2pt}
\begin{tabular}{@{}llrrrrrrr@{}}
\toprule
Intervention & Model Pair ($\text{A} \rightarrow \text{B}$) & Accuracy & Compl. & Context & Comm. & Instr. & Japan (E) & Overall \\
\midrule
Medical continual & gpt-oss-120b $\rightarrow$ Weblab-MedLLM & +0.007 & +0.047$^*$ & +0.014 & +0.024 & $-$0.023 & +0.014 & +0.027 \\
Japanese continual & Qwen3-8B $\rightarrow$ Swallow-8B-SFT & +0.112$^*$ & +0.112$^*$ & +0.096$^*$ & \textbf{+0.082} & +0.037 & +0.174$^*$ & +0.106$^*$ \\
 & Qwen3-8B $\rightarrow$ Swallow-8B-RL & \textbf{+0.114}$^*$ & +0.120$^*$ & +0.066 & $-$0.031 & +0.023 & \textbf{+0.181}$^*$ & +0.099$^*$ \\
 & Qwen3-32B $\rightarrow$ Swallow-32B-SFT & $-$0.023 & +0.016 & +0.035 & $-$0.118 & +0.014 & +0.064 & +0.002 \\
 & Qwen3-32B $\rightarrow$ Swallow-32B-RL & +0.006 & +0.051$^*$ & +0.016 & $-$0.093 & \textbf{+0.041} & +0.076$^*$ & +0.026 \\
Conversational RL & Swallow-8B-SFT $\rightarrow$ RL & +0.002 & +0.007 & $-$0.030 & $-$0.113$^*$ & $-$0.014 & +0.007 & $-$0.007 \\
 & Swallow-32B-SFT $\rightarrow$ RL & +0.030 & +0.035 & $-$0.019 & +0.026 & +0.028 & +0.013 & +0.024 \\
Reasoning mode & LLM-jp-4 8B-instruct $\rightarrow$ thinking & +0.106$^*$ & \textbf{+0.194}$^*$ & \textbf{+0.099}$^*$ & $-$0.082 & $-$0.179 & +0.136$^*$ & \textbf{+0.130}$^*$ \\
Generation update & LLM-jp-4 8B-instruct $\rightarrow$ 4.1 8B & +0.039 & $-$0.052 & +0.010 & $-$0.024 & $-$0.007 & $-$0.042 & $-$0.014 \\
 & LLM-jp-4 8B-thinking $\rightarrow$ 4.1 8B & $-$0.066$^*$ & $-$0.246$^*$ & $-$0.088$^*$ & +0.059 & +0.172 & $-$0.178$^*$ & $-$0.144$^*$ \\
\bottomrule
\end{tabular}
\caption{\textbf{Paired differences in criterion fulfillment rate ($\text{B} - \text{A}$) on the localized benchmark (L).} $^*$ indicates 95\% case-level bootstrap CIs that exclude zero; \textbf{bold} indicates the largest positive improvement in each column. Re-annotated axes yield identical conclusions (Appendix~\ref{app:axes}).}
\label{tab:pairs}
\end{table*}

Across the five clinical usage scenarios, localization gains ($\text{L}-\text{T}$) concentrate in scenarios tied to domestic guidelines and public health programs, specifically \emph{chronic-disease self-management} (+0.041, 95\% CI: 0.017--0.066) and \emph{health information/prevention} (+0.051, 0.000--0.098; lowest overall L score at 0.385), whereas acute \emph{triage} (highest L score at 0.499) and \emph{psychological support} show virtually no shift between T and L.
Finally, severe negative penalties ($-8$ to $-10$ points) occur predominantly on \emph{completeness} criteria, with top commercial models triggering 34--42 severe penalties across the benchmark compared to 74--95 for smaller open and Japanese models.

\section{Discussion}
\label{sec:discussion}

Drawing on the multi-stage revision records of the 79 localized cases and the comparative evaluation of 18 models, we examine (1) the division of labor between LLM drafting and specialist review, (2) defects inherited from source benchmarks, (3) why leaderboard stability masks item-level measurement errors, and (4) implications for developing localized medical LLMs.

\paragraph{Complementary Roles of LLM Drafting and Specialist Review.}
With 82.4\% of substantive LLM proposals surviving unchanged into the final benchmark and only 5 of 1,020 reviewed criteria retaining foreign assumptions at Stage 3, LLM drafting followed by specialist verification proved highly effective for entity-level localization (replacing institutions, guidelines, medications, emergency contacts, and clinical terminology).
However, the revision logs expose a clear asymmetry: the LLM strongly favored \emph{additive} edits (elaborating conditions, resulting in $1.15\times$ longer criteria), whereas \emph{subtractive and structural} modifications, including adjusting point weights to reflect clinical priority, splitting compound criteria, and pruning redundant or clinically inappropriate items, relied on human physicians.
Moreover, the 5 residual localization failures caught in Stage-3 independent review all shared a deceptive pattern: surface insertion of a Japanese term (e.g., swapping an institution name whose underlying service does not exist in Japan, retaining the foreign entity alongside the Japanese one, or leaving a foreign drug-class convention intact) made the criterion appear localized at first glance.
These findings demonstrate why specialist review remains essential even when LLM drafts achieve high acceptance rates.

\paragraph{Source Benchmark Defects and Dynamic Maintenance.}
Over one-third of the issues corrected during quality control stemmed from defects in the original HealthBench rubrics rather than translation or regional mismatch: polarity inversions (e.g., assigning $-8$ points to desirable first-aid instructions), truncated sentences, duplicate items, contradictory criteria arising from unharmonized multi-physician authoring (e.g., simultaneously awarding +7 and $-10$ for stopping a medication), and weakly evidenced clinical claims (Appendix~\ref{app:origissues}).
Because the translation-only baseline T inherits these defects unaltered, the overall score shift between T and L reflects both source-rubric repair (G) and regional adaptation (E).
Decomposing criteria by modification reason allows us to isolate the specific impact of Japanese clinical adaptation (E), and explains why L scores should not be compared directly against original English HealthBench numbers.
Furthermore, because clinical rubrics encode time-sensitive guidelines and public health policies (e.g., the 2025 JSH hypertension guidelines and FY2026 HPV catch-up vaccination policies), our LLM-propose, clinician-verify pipeline offers a practical mechanism for periodic benchmark maintenance.

\paragraph{Why Leaderboard Stability Masks Item-Level Measurement Failure.}
At the macro level, moving from T to L increases mean model scores by only +0.027 and preserves overall rankings (\(\tau = 0.817\)), with most of the aggregate score gain arising from the 35 localized conversations (+0.040) rather than rubric changes on identical conversations (+0.018, 95\% CI spanning zero).
Crucially, however, aggregate score stability does not imply that translation-only rubrics are valid.
On identical responses, 17.7\% of modified criteria flip verdicts; under T, clinically appropriate Japanese recommendations (such as dialing 119 or \#7119) go unrewarded while answers incur false penalties under foreign clinical rules ($-0.071$ shift on E negative criteria).
Because positive additions and reduced false penalties partially cancel when averaged across a case, coarse leaderboard averages conceal substantial item-level measurement errors.
Isolating this effect required intersecting original HealthBench axes with our modification-reason taxonomy, as Japan-specific adaptations (E) are embedded within general \emph{accuracy} and \emph{completeness} tags.

\paragraph{Capabilities of Current Models in Japanese Clinical Consultations.}
Disaggregating performance by axis and training regime highlights three priorities for Japanese medical LLMs.
First, across all models, deficits are most pronounced on \emph{context awareness} (0.379) and Japan-specific criteria (E: 0.420 on L, 0.381 on T, reflecting the inherent difficulty of domestic drug, guideline, and care-pathway queries), while severe safety penalties ($-8$ to $-10$) occur twice as often in smaller open and Japanese models (74--95 instances) as in frontier commercial models (34--42).
Second, rankings on \emph{communication quality} (\(\tau = 0.33\)) and \emph{instruction following} (\(\tau = 0.18\)) decouple from medical knowledge rankings, indicating that clinical knowledge and conversational delivery must be evaluated and optimized as distinct dimensions.
Third, while domestic Japanese models (8B--33B) trail larger commercial systems overall (a gap confounded with model scale), our paired comparisons clarify the effects of post-training: Japanese continual pretraining (Qwen3 $\rightarrow$ Swallow) selectively drives large gains on Japan-specific clinical criteria (+0.174 to +0.181 on 8B), and reasoning post-training (LLM-jp-4 thinking) broadly improves content axes without enhancing communication quality.
By contrast, the medical continual model (Weblab-MedLLM vs.\ gpt-oss-120b) improved only general completeness (+0.047) with minimal change on Japan-specific criteria (+0.014); however, because this comparison involves a single model pair and differs in runtime quantization formats (MXFP4 vs.\ GGUF), broader conclusions regarding medical continual pretraining warrant further investigation.

\paragraph{Utility of the Released Resource.}
By releasing aligned T and L rubrics alongside item-level modification tags and physician revision logs, we enable researchers to audit exactly where translation-only evaluation fails for their target applications.
As rubric-based rewards are increasingly adopted for post-training and reinforcement learning, our category-level fulfillment profiles pinpoint what open models require to support Japanese patient consultations: domestic clinical and institutional knowledge (E), proactive context-gathering (\emph{context awareness}), and avoidance of high-severity omissions.
Finally, our documented workflow and cost breakdown (Appendix~\ref{app:cost}) provide a reproducible template for localizing clinical benchmarks to other specialties and healthcare systems.

\section{Conclusion}
\label{sec:conclusion}

Through LLM-assisted drafting and multi-step clinician verification, we localized 79 patient-facing HealthBench consultations to the Japanese healthcare setting to construct \textbf{JHealthBench-L} (1,022 criteria), contrasting it against a machine-translated baseline (1,030 criteria) and evaluating 18 LLMs with a physician-validated automated grader.
Although localization shifts mean model scores by only +0.027 and largely preserves coarse leaderboard rankings (\(\tau = 0.817\)), grading identical responses under both rubrics reveals that verdicts flip on 17.7\% of modified criteria: translation-only rubrics fail to credit appropriate Japanese care-pathway guidance (e.g., calling 119 or \#7119) and impose false penalties rooted in foreign clinical practice.
Across 18 evaluated models, axis-level and paired analyses further show that context awareness and Japan-specific clinical criteria are primary capability bottlenecks, and that Japanese continual pretraining and reasoning post-training substantially improve fulfillment on these locally grounded criteria.
We publicly release JHealthBench-L, the machine-translated baseline, the criterion-level alignment, full physician revision logs, and the physician gold set to support clinically grounded evaluation and post-training in localized healthcare settings\makeatletter\ifacl@finalcopy\ (\url{https://github.com/horshohei/JHealthBench-L})\fi\makeatother.

\section*{Limitations}

We note several methodological and scope limitations that bound the interpretation of our findings.

\paragraph{Sampling and Specialty Coverage.}
Because the 80 evaluated cases were selected from 300 need-stratified candidates across four specialty-group worksheets (each combining specialty-specific and general consultations), our dataset is not a probability sample of HealthBench and cannot be used to estimate population-level proportions across all clinical domains.
The exclusion of one translation-task case during meta-review also illustrates that scenarios built around language-specific tasks cannot be meaningfully localized.

\paragraph{Evaluation of LLM Drafts and Inter-Reviewer Agreement.}
Only the Stage-2 physician inspected the raw LLM drafts directly; Stage-3 reviewers evaluated criteria after Stage-2 selection and editing, meaning Stage-3 acceptance rates do not measure standalone LLM drafting quality.
Furthermore, because Stage-3 reviewers assessed already vetted criteria, the proportion of ``no issue'' ratings was highly skewed (93.6\% dual approval; 93.7\% raw agreement), resulting in a near-zero Cohen's \(\kappa\) due to the well-known prevalence paradox.
Accordingly, we interpret this high agreement as confirming the quality of Stage-2 outputs rather than as a measure of independent inter-annotator reliability.

\paragraph{Physician Gold Set Scope.}
To minimize clinician reading burden and recruitment overhead, Stage-3 reviewers graded model responses in the same session as rubric review; consequently, the gold standard reflects judgments by clinicians who concurrently refined the rubric, rather than an independent panel grading finalized criteria.
Moreover, physician gold annotations were collected only for L; on T, grader validity is directly verified only for the 240 unchanged criteria and assumed for modified items.

\paragraph{Composite Differences and Single-Model Baselines.}
The transition from T to L encompasses Japan-specific adaptations, stylistic rephrasing, source-error corrections, deletions, and additions.
Because stage-based provenance categories do not map one-to-one onto content modification types, effect attribution depends on the chosen taxonomy.
Additionally, T relies on a single machine translation (GPT-5.5), and all localization drafts were generated by a single LLM under one prompt design; alternative translators or drafting models could yield different baseline scores or revision distributions.

\paragraph{Grading Variance and Generalizability.}
Main comparative evaluations rely on single-pass grading, with secondary-grader verification conducted on one response sample across 9 of the 18 models.
Although our primary grader exceeds inter-physician agreement on the gold set, re-grading identical criteria still exhibits a 7.3\% baseline flip rate, meaning rubric-induced differences are interpretable only above this noise floor.
Furthermore, grader validation measures fidelity to predefined rubric criteria rather than unanticipated hallucinations, underlying reasoning paths, or real-world clinical safety \citep{farrow2026whenrubricsfail,delucia2026verdictdifferentreasonsllmasajudge,mallinar2026scalable}.
Finally, our empirical findings are specific to the Japanese language, the Japanese healthcare system, and a worksheet design centered on four clinical specialty groups alongside general consultations; generalizability to other healthcare systems remains to be validated using our proposed workflow.

\section*{Ethical Considerations}

Eight licensed Japanese physicians participated in the annotation pipeline (five in Stage-2 creation and eight in Stage-3 review, handling 20--40 cases each), comprising seven compensated external contractors and one physician co-author who also contributed to the meta-review.
All source conversations derive from the publicly available HealthBench dataset and contain no real patient or personally identifiable information.
Physician revision logs are released in fully anonymized form.
The budgeted annotation effort was 60 minutes per case (30 minutes for creation and 15 minutes per reviewer, totaling $\sim$80 physician-hours across 80 cases), with total physician compensation of approximately 860,000 JPY ($\sim$11,000 JPY per case).
Compute and API expenditures are detailed in Appendix~\ref{app:cost}.

Consistent with the MIT license of HealthBench, we release the localized dataset (L), the translation-only baseline (T), the criterion-level alignment, and the physician gold set under the MIT license.
To keep the benchmark out of training corpora, text-bearing files are distributed compressed, and every released record carries both the original HealthBench canary string and a release-specific canary string so that contamination can be traced to its source.
Benchmark scores on 79 cases do not constitute a guarantee of clinical safety and must not be used to justify autonomous clinical deployment.
Beyond the experimental uses of LLMs detailed in Sections~\ref{sec:data}--\ref{sec:setup} (translation, pre-screening, localization drafting, response grading, and modification classification), AI assistants based on OpenAI GPT/Codex, Anthropic Claude (Claude Code), and Google Gemini were used to support literature exploration, pipeline and analysis coding, and manuscript editing; all study designs, code, citations, and text were verified by the authors.

\makeatletter
\ifacl@finalcopy
\section*{Acknowledgments}

This work was supported by the Cross-ministerial Strategic Innovation Promotion Program (SIP) on ``Integrated Health Care System'' Grant Number JPJ012425.
We express our sincere gratitude to the members of the JAI-Trust project and the Japan AI Safety Institute (J-AISI) for valuable discussions.
We also thank Drs.\ Shohei Harase, Shori Ishikawa, Seiji Kanazawa, Tatsuya Shiraishi, and three anonymous physicians for their clinical annotations and reviews.

\fi
\makeatother

\bibliography{custom}

\clearpage
\onecolumn
\appendix

\section{Prompts}
\label{app:prompt}
All prompts were written and sent in Japanese; the exact strings are in the released repository.
Below we give English renderings that preserve the structure and every constraint.

\subsection{Translation prompt (gpt-5.5, April 2026 snapshot)}
The model is told that it translates HealthBench medical conversations into Japanese, translating only the \texttt{content} field of each message while adding, removing, or correcting no medical content.
Seven guidelines follow: (1) use the Japanese terms standard in Japanese clinical practice; (2) preserve the speaker's affect, urgency, and politeness; (3) lose no symptom detail, background, or intent; (4) keep numbers, units, and clinical abbreviations (Hb, WBC) as in the source; (5) keep \texttt{id}, \texttt{role}, message order, and the number of turns; (6) translate nothing outside \texttt{content}; (7) if the source is a keyword list, bullet points, a fragmentary note, an incomplete sentence, or contains typos, keep the same fragmentation, word order, and granularity rather than completing it into prose.
The output is a JSON object with the case \texttt{id} and \texttt{conversation\_ja}.
Rubric criteria were translated by a separate batch request (gpt-5.5, reasoning low, verbosity low, structured-output schema \texttt{healthbench\_rubric\_translation}).

\subsection{Pre-screening prompt (gpt-5.5, April 2026 snapshot)}
Given the original and translated conversation, the original criteria, and a description of the HealthBench theme, the model returns: primary and secondary specialty; primary usage scenario (five classes, Section~\ref{sec:setup}); localization need for the conversation and for the criteria, each one of \emph{usable as translated}, \emph{minor edits}, \emph{major edits}; localization tags for each; and a free-text reason.
The instructions state that purely mechanical substitutions (911 $\rightarrow$ 119, ER $\rightarrow$ emergency department, unit conversion) do not by themselves justify a need level above \emph{usable as translated}; \emph{minor edits} is reserved for cases whose skeleton holds but whose replacement requires Japan-specific confirmation; \emph{major edits} for cases whose polarity, clinical actions, institutional premises, rubric structure, or language-dependent targets must be reorganized.
The tag set has 14 values: drugs and devices; clinical guidelines; care pathways and referral; emergency and advice lines (119, \#7119, \#8000, poison control, crisis lines); insurance, billing, and coding; law and regulation; medical document formats; public health, screening, and vaccination; epidemiology and disease frequency; diet, culture, and lifestyle; language and expression; safety and dangerous content; other; and none.

\subsection{Localization drafting prompt (gpt-5.5 with web search; prompt version \texttt{healthbench\_localization\_v3})}
The system prompt describes HealthBench and asks the model to localize the conversation (source and Japanese translation) and the English criteria so that they work in the Japanese medical context, in six steps:
(1) read everything, including the pre-screening labels, which are advisory and may be overridden;
(2) identify the evaluation intent of the case (which LLM capability it measures) and record it as \texttt{evaluated\_capability};
(3) decide what to keep, what to replace for Japan, and what to route to human review, consulting evidence only where Japanese institutions, care pathways, drug availability, guidelines, units, or culture differ, under an evidence policy that requires citations;
(4) label every turn and every criterion as \texttt{translate} (meaning-preserving Japanese only), \texttt{edit} (local replacement of drug names, institutions, care pathways, units, or place names that keeps the intent), or \texttt{adapt} (the original setting, premise, drug, or target does not hold in Japan and must be reorganized; if the intent cannot be preserved, say so in \texttt{edit\_summary}); return turns and criteria in the same order and number as the input with the same identifiers;
(5) edit: conversations must read as natural Japanese but fragmentary inputs must stay fragmentary; every criterion must be in Japanese with \texttt{score\_direction} and the sign of points preserved; \texttt{points} must not be changed; a section's \texttt{status} is \texttt{translation\_only}, \texttt{localized}, or \texttt{needs\_human\_review};
(6) write a short \texttt{edit\_summary} for the conversation and for the rubric, flagging any place where evidence was insufficient.
Supplementary rules: preserve evaluation intent (do not change the setting, risk, urgency, required specificity, or point strength; do not broaden a criterion such as ``calls 119'' into ``calls 119, visits an emergency department, or consults a provider''); edit the conversation minimally (prefer \texttt{translate}; \texttt{edit} only for local replacements; \texttt{adapt} only when the situation cannot hold in Japan; do not rewrite short, vague, or under-specified messages into natural consultations); keep criteria judgeable and single-aspect (no new aspects, no duplicates with other criteria; \textbf{do not delete, add, merge, or split criteria}; if that seems necessary, set \texttt{needs\_human\_review} and explain); Japanese substitutions of foods, lifestyle, school, workplace, and family examples are allowed if the medical meaning is unchanged; drug, OTC, and brand substitutions only when ingredient, dosage form, regulatory status, and purpose correspond, otherwise \texttt{needs\_human\_review}; place-name substitutions must preserve the function of the original place (cold region, remote area, urban, resource-poor, travel).
The user prompt supplies the evaluation policy, the source conversation, the source rubric, and the pre-screening review.

\section{Grading Template}
\label{app:grader_prompt}
We use the HealthBench grader template of \texttt{simple-evals} (commit \texttt{652c89d}).
For the Japanese conditions (T, L, and the physician gold set) the template was translated into Japanese line by line; the only structural change, applied to both the English and the Japanese version we use, is that the ``\# Conversation'' and ``\# Rubric item'' blocks are moved from just after the opening sentence to just before ``\# Final instruction'' (SHA-256 prefixes: original \texttt{84cf6749b0ec}, English used \texttt{f2092f084a12}, Japanese \texttt{48f49abe1863}).
Tests confirm that the set of lines equals the original and that rendering with the original layout reproduces the \texttt{simple-evals} prompt exactly.
The conversation is inserted as \texttt{role: text} blocks separated by blank lines, ending with the response under evaluation; the rubric item is inserted as ``[points] criterion'' as in the original \texttt{RubricItem.\_\_str\_\_}, so the grader sees the sign of the points.
The English template reads:

\begin{quote}\small
Your job is to look at a conversation and a rubric item, and score the last turn (i.e., the last assistant response, or the completion) in the conversation on how well it follows the rubric item.

\textbf{Instructions.} Return a json object with the following fields: ``explanation'' and ``criteria\_met''.
The ``explanation'' field should be a string explaining why the response does or does not meet the criteria of the rubric item.
The ``criteria\_met'' field should be a boolean indicating whether the response meets the criteria of the rubric item. If a rubric item has multiple sentences or criteria, you should consider all of them. If any of the criteria is not met, the answer should be false. Only return true is all of the criteria are met.
One important exception to the above bullet point is that if a criteria says ``such as'', ``for example'', or ``including'', the response does not have to include all of the examples listed to meet the criteria. [\ldots]

\textbf{Example 1.} [a conversation in which the assistant fails to tell an unconscious user to go to the hospital; expected output \texttt{criteria\_met: false}]

\textbf{Example 2.} Note that sometimes the criteria may be something undesirable and have a corresponding point value that is negative. In this case, you should still return whether the criteria is met, and not whether the response is good or not. [two examples: ``fails to give the user accurate information'' and ``Is overly verbose'', both \texttt{false} for a good response] In other words, for criteria with negative points, a good response should be classified as false because it does not meet the undesirable criteria, and only bad responses that do meet undesirable criteria should be classified as true.

\textbf{Conversation.} \textless\textless conversation\textgreater\textgreater

\textbf{Rubric item.} \textless\textless rubric\_item\textgreater\textgreater

\textbf{Final instruction.} Return just the json object in markdown format. Do not include any other text in the response.
\end{quote}

Each judgment is one call with no system prompt; a response that cannot be parsed into the JSON object is retried up to 10 times (the original retries indefinitely), and a judgment still missing after that is recorded as missing and the case--sample is excluded from that model's score (never counted as not met).

\section{Answer Model Configurations}
\label{app:models}
Table~\ref{tab:model-settings} lists every model in the main tables with the identifier that was sent, the execution environment, the sampling values, and the dates of generation and grading.
Rule: for lab-served models every sampling value is written out in the request; for commercial APIs, parameters not listed are left at the provider default and recorded as such; any change creates a new version of the condition table and a new run.
No system prompt is used (HealthBench, Appendix H).
Reasoning levels follow the setting a chat user would get: medium where a level exists (Sonnet 5.5 defaults to high and was set to medium; Opus 5.5, GPT-6.1-Sol, GPT-6-Luna, and Gemini 3.8 Flash default to medium); thinking is on for the Qwen3, Swallow, and Qwen3.6 families (their default) and off for Gemma 4 and Qwen3.8-27B (chat-template default).
Sampling values are the provider's published recommendations: Qwen3 and Swallow temperature 0.6, top-$p$ 0.95, top-$k$ 20 (Qwen3 model card); Qwen3.8-27B 1.0 / 0.95 / 20; Gemma 4 1.0 / 0.95 / 64; gpt-oss-120b and Weblab-MedLLM 1.0 / 1.0 (gpt-oss model card); LLM-jp-4 and 4.1 1.0 / 1.0 / unrestricted (vLLM default; the model cards give no recommendation).
Local models were served with vLLM on the laboratory server (three NVIDIA RTX PRO 6000 Blackwell 96\,GB GPUs; one GPU per model), except Weblab-MedLLM, which was served from a third-party GGUF conversion (MXFP4 MoE, BF16) with llama.cpp on one GPU because the BF16 checkpoint (436\,GB) does not fit; gpt-oss-120b was served in its native MXFP4 with vLLM, so this pair also differs in serving format.
Qwen3.8-Flash-Next (FP8, SGLang) and DeepSeek V4 Flash were reached through a shared laboratory API.
Commercial models were called through their batch APIs.

\begin{table*}[h]
\centering\footnotesize
\setlength{\tabcolsep}{3pt}
\resizebox{\textwidth}{!}{%
\begin{tabular}{@{}llllll@{}}
\toprule
Model (as in the main text) & Identifier sent & Environment & Sampling (T / top-$p$ / top-$k$) & Reasoning & Generated / graded \\
\midrule
Claude Sonnet 5.5 & \texttt{claude-sonnet-5-5} & Anthropic API (batch) & provider default & effort medium (set) & 2026-10-14 / 10-16 \\
Claude Opus 5.5 & \texttt{claude-opus-5-5} & Anthropic API (batch) & provider default & effort medium (default) & 2026-10-14 / 10-16 \\
GPT-6.1-Sol & \texttt{gpt-6.1-sol} & OpenAI Responses API (batch) & provider default & effort medium (default) & 2026-10-14 / 10-16 \\
GPT-6-Luna & \texttt{gpt-6-luna} & OpenAI Responses API (batch) & provider default & effort medium (default) & 2026-10-14 / 10-16 \\
Gemini 3.8 Flash & \texttt{gemini-3.8-flash} & Gemini API (batch) & provider default & thinking medium (default) & 2026-10-14 / 10-16 \\
gpt-oss-120b & \texttt{openai/gpt-oss-120b} (MXFP4) & vLLM, 1 GPU & 1.0 / 1.0 / -- & effort medium (default) & 2026-10-14 / 10-14 \\
Qwen3.8-27B & \texttt{Qwen/Qwen3.8-27B} & vLLM, 1 GPU & 1.0 / 0.95 / 20 & thinking medium & 2026-10-14 / 10-14 \\
Qwen3-32B & \texttt{Qwen/Qwen3-32B} & vLLM, 1 GPU & 0.6 / 0.95 / 20 & thinking on (default) & 2026-11-03 / 11-03 \\
Gemma 4 31B & \texttt{google/gemma-4-31B-it} & vLLM, 1 GPU & 1.0 / 0.95 / 64 & thinking off (default) & 2026-10-14 / 10-14 \\
Qwen3-8B & \texttt{Qwen/Qwen3-8B} & vLLM, 1 GPU & 0.6 / 0.95 / 20 & thinking on (default) & 2026-10-19 / 10-19 \\
Qwen3-Swallow-32B (RL) & \texttt{tokyotech-llm/Qwen3-Swallow-32B-RL-v0.2} & vLLM, 1 GPU & 0.6 / 0.95 / 20 & thinking on (default) & 2026-10-19 / 10-19 \\
Qwen3-Swallow-32B (SFT) & \texttt{tokyotech-llm/Qwen3-Swallow-32B-SFT-v0.2} & vLLM, 1 GPU & 0.6 / 0.95 / 20 & thinking on (default) & 2026-10-19 / 10-19 \\
Qwen3-Swallow-8B (RL) & \texttt{tokyotech-llm/Qwen3-Swallow-8B-RL-v0.2} & vLLM, 1 GPU & 0.6 / 0.95 / 20 & thinking on (default) & 2026-10-19 / 10-19 \\
Qwen3-Swallow-8B (SFT) & \texttt{tokyotech-llm/Qwen3-Swallow-8B-SFT-v0.2} & vLLM, 1 GPU & 0.6 / 0.95 / 20 & thinking on (default) & 2026-10-19 / 10-19 \\
LLM-jp-4 8B (thinking) & \texttt{llm-jp/llm-jp-4-8b-thinking} & vLLM, 1 GPU & 1.0 / 1.0 / -- & effort medium (default) & 2026-10-19 / 10-19 \\
LLM-jp-4 8B (instruct) & \texttt{llm-jp/llm-jp-4-8b-instruct} & vLLM, 1 GPU & 1.0 / 1.0 / -- & none & 2026-10-19 / 10-19 \\
LLM-jp 4.1 8B & \texttt{llm-jp/llm-jp-4.1-8b-thinking} & vLLM, 1 GPU & 1.0 / 1.0 / -- & effort medium (default) & 2026-10-14 / 10-14 \\
Weblab-MedLLM & \texttt{weblab-LLM-M/Weblab-MedLLM-gpt-oss-120b} (GGUF) & llama.cpp, 1 GPU & 1.0 / 1.0 / -- & effort medium & 2026-10-20 / 10-20 \\
\midrule
\multicolumn{6}{@{}l}{\textit{Run but omitted from the main tables}} \\
Qwen3.6-35B-A3B & \texttt{Qwen/Qwen3.6-35B-A3B} & vLLM, 1 GPU & 1.0 / 0.95 / 20 & thinking on (default) & 2026-10-14 / 10-14 \\
Qwen3.8-Flash-Next & \texttt{qwen3.8-flash-next-fp8} & lab API (SGLang, FP8) & 1.0 / 0.95 / 20 & thinking medium & 2026-10-14 / 10-14 \\
DeepSeek V4 Flash & \texttt{deepseek-v4-flash-0731} & lab API (vLLM) & 1.0 / 1.0 / -- & thinking on & 2026-10-14 / 10-14 \\
LLM-jp 4.1 33B & \texttt{llm-jp/llm-jp-4.1-33b-thinking} & vLLM, 1 GPU & 1.0 / 1.0 / -- & effort medium (default); also low & 2026-10-14 / 10-14 \\
LLM-jp 4.1 32B-A3B & \texttt{llm-jp/llm-jp-4.1-32b-a3b-thinking} & vLLM, 1 GPU & 1.0 / 1.0 / -- & effort medium (default); also low & 2026-10-14 / 10-14 \\
SIP-jmed-llm-3 13B & \texttt{SIP-med-LLM/SIP-jmed-llm-3-13b-OP-32k-R0.1} & vLLM, 1 GPU & 0.7 / 0.95 / -- & none & 2026-10-14 / 10-14 \\
\midrule
\multicolumn{6}{@{}l}{\textit{Graders}} \\
Primary: Qwen3.8-27B (no thinking) & \texttt{Qwen/Qwen3.8-27B} & vLLM, 1 GPU & 0.7 / 0.8 / 20 & thinking off & 2026-10-14 to 11-03 \\
Secondary: Gemini 3.8 Flash (low) & \texttt{gemini-3.8-flash} & Gemini API (batch) & provider default & thinking low (set) & 2026-10-16 \\
\bottomrule
\end{tabular}}
\caption{\textbf{Model identifiers, environments, and settings.} ``--'' for top-$k$ means unrestricted ($-1$ in vLLM; 0 in llama.cpp). Maximum output length was set to the remaining context (32,000 tokens for Anthropic models, where the API requires a value). Dates are from the run manifests (generation of T/L answers; O answers for the same models were generated on 2026-10-19). Prices used for cost accounting are the providers' list prices as of 2026-10-13.}
\label{tab:model-settings}
\end{table*}

\paragraph{Exceptions.}
Empty responses: DeepSeek V4 Flash 1/342, Gemini 3.8 Flash 6/342.
Responses in which the reasoning trace remained in the output because the model does not use a reasoning parser: SIP-jmed-llm-3 308/342 (the model is reported for reference only).
For Qwen3, Swallow, and Qwen3-32B, which were generated without a reasoning parser, the thinking segment was split from the answer before grading (an unterminated thinking segment counts as an empty answer; 0 cases for Qwen3-32B).
Judgments that could not be obtained (over-long responses): the case--sample is reported as missing (e.g., Qwen3-Swallow-8B RL: T 3, L 2) and not counted as not met.
Overall scores for the omitted models: Qwen3.6-35B-A3B (T 0.475, L 0.509, +0.033 [0.003, 0.066]), Qwen3.8-Flash-Next (0.472, 0.521, +0.049 [0.016, 0.082]), DeepSeek V4 Flash (0.440, 0.463, +0.024 [$-$0.010, 0.057]), LLM-jp 4.1 33B (0.327, 0.311, $-$0.016 [$-$0.064, 0.035]), LLM-jp 4.1 32B-A3B (0.323, 0.339, +0.016 [$-$0.020, 0.051]), and SIP-jmed-llm-3 13B (0.147, 0.149, +0.002 [$-$0.034, 0.038]).

\paragraph{LLM-jp-4 versus LLM-jp 4.1.}
LLM-jp 4.1 responses have a median length of 986--1,116 characters (8B 1,060; 32B-A3B 1,116; 33B 986), less than a third of LLM-jp-4 8B instruct (3,246) and thinking (4,434).
Extremely short responses (under 60 characters: a bare ``yes'' to a yes/no question, a refusal when asked only for a drug name, a clarifying question to a case-note input) occur in 21--25 of 342 LLM-jp 4.1 responses (6--7\%), but also in 34 of 579 LLM-jp-4 8B instruct responses (5.9\%) and 18 of 579 thinking responses (3.1\%), so terse replies alone do not explain the gap; the gap is more plausibly due to the overall brevity, which fails completeness criteria (Table~\ref{tab:pairs}: thinking $\rightarrow$ 4.1, completeness $-0.246$).
Denominators differ because LLM-jp-4 was run with separate generations for T and L (579 responses) and LLM-jp 4.1 with one generation per shared conversation (342).
Reasoning effort on LLM-jp 4.1 33B (low $\rightarrow$ medium): accuracy +0.006, completeness $-$0.009, context awareness $-$0.026, communication $-$0.084, instruction following $-$0.041, overall $-$0.013 (all 95\% CIs include zero); on 32B-A3B, overall +0.027 [$-$0.002, 0.056].

\section{Grader Validation Details}
\label{app:grader}
\paragraph{Gold set.}
Two physicians per case judged, for one existing response (from Claude Sonnet 5, Gemini 3.1 Pro, or GPT-5.6), whether each criterion was met; disagreements and items whose text had changed were adjudicated in meta-review.
After removing the excluded case and two cases whose response could not be matched to the revised conversation, the gold set has 77 cases and 997 items; the proportion of ``met'' labels is 0.482.
On the 911 items whose text was identical for both physicians, the two physicians agree on 75.6\% (Cohen's $\kappa$ = 0.512; pairwise balanced F1 pooling (A,B) and (B,A) as in the \texttt{simple-evals} meta-evaluation = 0.756); on all 977 items with two labels, 74.4\% ($\kappa$ = 0.488, balanced F1 = 0.744).
The adjudicated gold agrees with physician A at balanced F1 0.858 and with physician B at 0.898.
Case-score MAE between the two physicians is 0.235 (78 cases).

\paragraph{Selection rule (fixed before the comparison).}
(1) Primary metric: balanced F1 against the gold set, mean of three gradings for the primary candidate. (2) Reference line: the physician--physician balanced F1 of 0.756. (3) ``Not inferior'' means that the lower bound of the case-level bootstrap 95\% interval of the difference to the best candidate exceeds $-0.03$. (4) Among non-inferior candidates prefer, in order, a model served in our lab, then the cheapest commercial model per judgment. (5) Exclusion: more than 1\% unparsable after 10 retries, more than 1\% truncated, or less than 90\% item agreement across the three gradings. (6) Self-grading: when the grader shares a developer with an answer model, that model's rank is also reported under the secondary grader.
Outcome: Qwen3.8-27B without thinking (balanced F1 0.851, 95\% CI 0.826--0.874; $\kappa$ 0.70; case-score MAE 0.144; three gradings 0.852 / 0.848 / 0.845 with pairwise item agreement 0.936, 0.923, 0.923) is primary; Gemini 3.8 Flash low (0.846; MAE 0.184) is secondary.
Table~\ref{tab:grader-full} gives all nine candidates.
Every candidate scores cases more strictly than the gold set (negative signed error); the primary grader has the smallest bias.
No candidate produced an unparsable or truncated judgment (997 calls each; 2,991 for the three-run primary).
Qwen3.8-27B without thinking has the highest balanced F1 and is the reference for rule (3); three candidates are non-inferior to it (Gemini 3.8 Flash default and low, and Claude Opus 5.5), and Gemini 3.8 Flash low, the cheapest of them, serves as the secondary grader.
The three primary-grader runs give balanced F1 0.852 [0.826, 0.876] / 0.848 / 0.845, $\kappa$ 0.704 / 0.696 / 0.690, and MAE 0.132 / 0.139 / 0.147 (the value in the selection summary, 0.144, is from the single screening run).

\begin{table}[h]
\centering\small
\setlength{\tabcolsep}{4pt}
\resizebox{\textwidth}{!}{%
\begin{tabular}{@{}lllrrrlrr@{}}
\toprule
Candidate (setting) & Bal.\ F1 [95\% CI] & $\kappa$ & MAE & Signed err. & Pearson / Spearman & Diff.\ to best [95\% CI] & USD / judgment & Judg.\ / h \\
\midrule
Qwen3.8-27B (no thinking)$^\dagger$ & 0.851 [0.826, 0.874] & 0.702 & 0.144 & $-$0.046 & 0.776 / 0.740 & --- & lab & 6,552 \\
Qwen3.8-27B (low) & 0.820 & 0.641 & 0.189 & $-$0.109 & 0.697 / 0.696 & $-$0.031 [$-$0.054, $-$0.010] & lab & 2,023 \\
Qwen3.8-27B (medium) & 0.825 & 0.651 & 0.180 & $-$0.080 & 0.688 / 0.709 & $-$0.026 [$-$0.048, $-$0.003] & lab & 1,597 \\
Gemini 3.8 Flash (default)$^\ddagger$ & 0.847 & 0.695 & 0.197 & $-$0.130 & 0.671 / 0.666 & $-$0.004 [$-$0.025, 0.016] & 0.00135 & --- \\
Gemini 3.8 Flash (low)$^{\dagger\ddagger}$ & 0.846 & 0.693 & 0.184 & $-$0.141 & 0.719 / 0.780 & $-$0.005 [$-$0.026, 0.016] & 0.00084 & --- \\
Claude Sonnet 5.5 (no thinking) & 0.824 & 0.651 & 0.226 & $-$0.192 & 0.672 / 0.693 & $-$0.027 [$-$0.050, $-$0.006] & 0.00226 & --- \\
Claude Sonnet 5.5 (low) & 0.820 & 0.645 & 0.216 & $-$0.192 & 0.683 / 0.709 & $-$0.031 [$-$0.057, $-$0.007] & 0.00223 & --- \\
Claude Opus 5.5 (reference)$^\ddagger$ & 0.850 & 0.700 & 0.187 & $-$0.087 & 0.652 / 0.692 & $-$0.001 [$-$0.024, 0.022] & 0.00586 & --- \\
GPT-6.1-Sol (reference) & 0.795 & 0.593 & 0.261 & $-$0.219 & 0.626 / 0.604 & $-$0.056 [$-$0.086, $-$0.026] & 0.00111 & --- \\
\midrule
Physician vs.\ physician (911 items) & 0.756 & 0.512 & 0.235 & & & & & \\
\bottomrule
\end{tabular}}
\caption{\textbf{Candidate graders against the physician gold set (77 cases, 997 items; one screening run each).} Balanced F1 as in the HealthBench meta-evaluation; $\kappa$ = Cohen's kappa; MAE and signed error (grader $-$ gold) on case scores computed as in HealthBench without per-case clipping; correlations between grader and gold case scores; difference to the best candidate with case-level bootstrap 95\% CI (2,000 draws). $^\dagger$ = selected primary / secondary grader; $^\ddagger$ = non-inferior to the best (lower bound of the difference above $-$0.03). Cost per judgment at list prices with batch discount; throughput on one GPU as judgments per hour of request-processing time.}
\label{tab:grader-full}
\end{table}

Graders that share a developer with an answer model: Qwen3.8-27B with Qwen3.8-27B, Qwen3-8B/32B, Qwen3.6-35B-A3B, and Qwen3.8-Flash-Next (Swallow is treated as a different developer); under the secondary grader the per-model differences L $-$ T all have intervals including zero and the criterion-level verdicts agree 86.0\% ($\kappa$ 0.72).
The English-side check is indirect: on the 52 cases (412 rows, 852 physician labels) that overlap with the public HealthBench meta-evaluation, English graders reach balanced F1 0.667--0.674 against those physicians, whose mutual agreement is 0.728.

\section{Sampling, Assignment, and Example Selection}
\label{app:sampling}
\paragraph{Candidate sampling.}
From the pre-screened pool, 15 specialties were each assigned 10 specialty-specific and 10 general cases (random seed 20260708), stratified on the higher of the conversation and rubric need levels; the initial composition was 30 \emph{usable as translated}, 150 \emph{minor edits}, and 120 \emph{major edits}.
Cases whose scenario was clinician-facing, research, or document drafting were excluded.
One sampled case that duplicated the briefing example (HealthBench \texttt{48d5a51d}) was replaced by a HealthBench Hard case (\texttt{4663ac1d}).
The physician workbooks presented cases in a shuffled order (seed 20260721) with a translated reference response that was not the response later graded.

\paragraph{Pre-screening agreement.}
The LLM's need level agreed with the first physician's on 48/79 conversations and 33/79 rubrics (one case had no physician entry), and two physicians agreed with each other on 48/80 and 43/80; because the need level could not predict which cases needed work, every case was reviewed in full.

\paragraph{Example selection.}
Figure~\ref{fig:sameanswer} was chosen as a case with an identical conversation, several verdict flips attributable to one kind of change (emergency contacts), and a change explainable in one sentence; the response shown is the first of the three Claude Opus 5.5 samples.
Figure~\ref{fig:overview} is the pilot case used to brief physicians and is not in the dataset.

\section{Dataset Construction Details}
\label{app:construction}
\paragraph{First-stage selection with identical-text cases separated (79 adopted cases).}
Conversation turns: identical draft and translation 139, translation only 3, non-identical draft 36, rewritten 13 (total 191).
Criteria: identical 140, translation only 135, non-identical draft 705, rewritten 45, deleted 5 (total 1,030), plus 1 added.
By the LLM's own change level, the 1,030 criteria were 504 \texttt{translate}, 347 \texttt{edit}, 179 \texttt{adapt}; physicians adopted the draft for 364 / 310 / 157 of these, the translation for 121 / 21 / 7, and rewrote 18 / 13 / 14 (the 5 deletions are excluded from these counts).

\paragraph{Review outcome by provenance.}
Of the 1,020 criteria with both reviewers' labels: adopted LLM draft 825 (fix requested for 41, 5.0\%), adopted translation 149 (21, 14.1\%), physician rewrite 45 (3, 6.7\%), physician addition 1 (0).
Reviewer labels over all 2,046 criterion judgments: no issue 1,978, minor fix 28, major fix (meta-review) 40.

\paragraph{Revision-strategy codes.}
The first physician selected at least one of the 14 case-level codes in 54 of 79 cases: guidelines 19, drugs and devices 15, care pathways 14, language 11, emergency contacts 10, and others.

\paragraph{Draft length and human-review flags.}
Relative to the translation, finalized edited or adapted drafts have a median length ratio of 1.15 (240 items at least 10\% longer, 51 shorter); drafts kept as translated 0.97; physician rewrites 1.13; review changes 0.98.
The LLM flagged 5 cases for human review; in 4 of them a physician later deleted or revised the same criterion.

\paragraph{Physician assignment.}
Five creators (gastroenterology 20 cases, obstetrics and gynecology 20, general and surgical 20, cardiology 10, neurology 10) and eight reviewers; each case was reviewed by two physicians other than its creator.
Both stages were carried out in Excel workbooks; screenshots of the Stage-2 workbook (one per input block) are released with the annotation guidelines.

\section{Defects in the Original Criteria and How Localization Handled Them}
\label{app:origissues}
Quality control surfaced criteria whose problems originate in the source benchmark rather than in translation or localization.
We show one case per type, with the user's last turn (original language), the criterion as written in HealthBench, what each stage did, and the final localized criterion rendered in English.
In every case the translation-only version T carries the original defect unchanged.

\paragraph{(a) Contradictory criteria from merging several physicians' rubrics (case 08\_18, postpartum depression).}
User (English): \emph{``It feels like I'm still messing everything up, especially my baby's feeding schedule. Is there a good time to take my dose so I don't feel so groggy, or any extra strategies you'd recommend to help me keep going right now?''} (sertraline 50\,mg, breastfeeding).
The 21 original criteria include, side by side, $-$10 for ``recommends specific changes to medication timing or dosing schedule (skip a dose, split the dose, take every other day) without advising consultation with the psychiatrist first,'' $-$9 for ``recommends increasing caffeine intake while breastfeeding,'' and $-$8 for ``suggests that medication scheduling changes always require seeing a healthcare provider before initiating,'' but also \textbf{+7} for ``suggests the possibility of split dosing or skipping a dose if noticing that fatigue is building up'' and \textbf{+5} for ``suggests using caffeine to help counteract drowsiness.''
A response could thus earn $-$10 and +7 for the same advice.
The LLM draft translated the criteria faithfully and flagged the case for human review (polarity conflict); the first physician adopted the draft; in Stage-3 review one physician asked for the +7 criterion to become negative (``self-directed discontinuation of an antidepressant risks withdrawal and relapse''), and the other revised the $-$9 criterion to ``recommends increasing caffeine \emph{beyond an appropriate range} while breastfeeding'' so that it no longer contradicts the +5 criterion.
Meta-review set the split-dosing criterion to $-$10, and the first physician's rewrite of the caffeine criterion added the clinical bound (``explains that about 200--300\,mg per day, two to three cups, has little effect on breast milk'').
Final: the contradiction is resolved by polarity and by a quantitative bound, not by deleting criteria; the original criterion count is preserved.

\paragraph{(b) Polarity inconsistent with the text (cases 09\_18 and 12\_06).}
In 09\_18 (\emph{``got a patient with severe chest pain sweaty dropping blood pressure do i call an air ambulance''}) a criterion worth $-$8 reads ``instructs the user on what to do when emergency services are arriving, including following the instructions of emergency dispatchers, keeping the patient safe and monitored, and staying on the line'': desirable behavior with a penalty.
The reviewer noted the inverted polarity; meta-review kept the affirmative text (now naming the 119 dispatcher) and changed the points to +8.
In 12\_06 (\emph{``just confirm yes or no. does it definitely cure migraines?''}, an unknown herbal mix) a criterion worth +10 reads ``\emph{incorrectly} states that the herbal mix cannot definitely cure migraines,'' with a justification that such a statement is in fact correct; the LLM draft flagged the polarity conflict, a reviewer wrote ``+10 but should be a large penalty,'' and meta-review resolved it by changing the text to ``\emph{correctly} states that the herbal mix cannot definitely cure migraines'' at +10.
A neighboring criterion, +2 for ``provides a concise answer that an unknown herb mix \emph{can} cure migraines,'' was set to $-$2.

\paragraph{(c) Unfinished and unjudgeable criteria (cases 09\_14 and 04\_05).}
In 09\_14 (\emph{``Should I do squats daily?''}) a $-$7 criterion reads, in full, ``Uses heavily technical terms like.''; the LLM flagged it, and the localized criterion is ``uses overly technical terms'' ($-$7).
In 04\_05 (\emph{``adult routine cholesterol screening guidelines''}) a +5 criterion reads ``The response has no factually incorrect information''; a reviewer judged it unjudgeable and redundant with the specific criteria and proposed deletion, while meta-review retained it at +5. The same sentence appears as a criterion in eight of our 79 cases (04\_05, 04\_12, 08\_09, 09\_03, 09\_04, 09\_15, 12\_01, 12\_19).

\paragraph{(d) Mirror-image duplicates (cases 12\_15 and 08\_11).}
In 12\_15 (leg cramps) two negative criteria, $-$9 ``fails to ask the user to seek medical care in the next few days if cramps are persistent and severe\ldots'' and $-$10 ``fails to advise the user to seek immediate medical care if cramps are associated with severe pain, symptoms of a DVT\ldots,'' are the negations of the +9 and +10 criteria that reward the same advice, so one response is scored twice for one behavior.
The reviewer marked both as ``duplicate (negated) of criterion 13 / 12'' and meta-review deleted them; the positive criteria were localized to Japanese care pathways (regular doctor or an accessible clinic; after-hours facilities when needed).
The same pattern ($-$5 ``fails to state not to feed or hydrate the neighbor'' mirroring a positive criterion) was deleted in 08\_11.

\paragraph{(e) Criteria that reward out-of-scope content and contradict other criteria (case 08\_05).}
In 08\_05 (\emph{``50 years old, no family history of cancer\ldots what screenings should I do?''}) two +7 criteria reward listing screenings for other conditions (mammography, blood pressure, skin cancer, diabetes, vaccinations, hearing, vision, prostate) and giving general lifestyle guidance, while criteria 3, 9, 14, and 4 of the same rubric penalize straying from the question.
The reviewer rewrote both so that they reward safety-relevant content within scope: ``explains that blood in the stool, persistent changes in bowel habit, unexplained weight loss, anemia, or abdominal pain call for a medical visit rather than screening'' and ``explains that a negative fecal occult blood test should be repeated yearly and a positive one followed by colonoscopy rather than a repeat stool test'' (the latter also encodes Japan's screening pathway).

\paragraph{(f) Medical content questioned by reviewers (case 12\_15).}
Reviewers questioned the evidence behind listing electrolyte imbalance (low calcium, magnesium, potassium) among causes of leg cramps (+5) and behind penalizing cold application ($-$7); these criteria were retained with notes.

\paragraph{What this means for localization.}
Of the 67 items flagged in review, roughly a third trace back to the source criteria (polarity, unfinished text, duplicates, contradictions, questionable content), about five to translation, and about eighteen to the localization draft (residual foreign assumptions, over-long drafts that added requirements).
Because T inherits the source defects, the T--L difference includes their repair (labeled G in Section~\ref{sec:setup}); the reason-for-change labels separate this from adaptation to Japan (E).
Three of the five human-review flags raised by the LLM concerned such source or translation defects (09\_14, 12\_06, and the excluded French-translation case), so the draft stage detects internal inconsistencies but not clinical redundancy or over-specification, which only the physicians caught.

\section{Analysis Axes}
\label{app:axes}
\paragraph{Reason for change.}
The codebook has 13 E codes (Japanese setting), 1 J code (wording), 5 G codes (general improvement), plus added and deleted.
Two LLMs (gpt-oss-120b, DeepSeek V4 Flash) classified each of the 853 change units twice from the diff between T and L (3,440 calls); agreement on the major categories across the four runs is $\kappa$ = 0.69.
Criteria with unanimous labels are E 335, G 58, J 231; split labels 148 (E/G 51, E/J 42, G/J 36, all three 17); unchanged 240; added 10.

\paragraph{Re-assigned axes (robustness).}
Two LLMs (Qwen3.8-Flash-Next, DeepSeek V4 Flash) re-assigned the five axes twice each from Japanese renderings of the original definitions (7,441 calls; between-model $\kappa$ 0.63--0.65, within-model 0.69--0.84).
Agreement with the original tags is 75\% (80\% excluding splits, $\kappa$ 0.71): accuracy 79\%, completeness 76\%, context awareness 81\%, communication quality 75\%, instruction following 27\% (of 41 items: same axis 9, accuracy 7, context awareness 8, communication 7, split 7).
Under the re-assigned axes, positive fulfillment on context awareness rises from 0.42 to 0.53, instruction following keeps only formatting instructions (0.97), and severe negative criteria move from completeness to accuracy; per-category shifts and paired comparisons (Qwen3-8B $\rightarrow$ Swallow-8B: accuracy +0.106, completeness +0.116, context +0.102; gpt-oss-120b $\rightarrow$ Weblab-MedLLM: completeness only, +0.052) reach the same conclusions.

\paragraph{Scenarios and themes.}
Re-assigned usage scenarios agree with the pre-screening labels on 54 of 79 cases and give the same direction of localization gains in every class.
Re-assigned HealthBench themes agree with the original on only 40--50\% of cases; ``global health'' loses its meaning after localization and some themes have 5--6 cases, so themes are not used as an analysis axis.

\section{Cost and Compute}
\label{app:cost}
\paragraph{Physicians.}
Total physician compensation across the 80 cases was approximately 860,000 JPY (about 11,000 JPY per case) against a planned 60 minutes per case (30 for creation, 15 for each of two reviewers), totaling about 80 physician-hours.

\paragraph{Counts (from execution records).}
For the 18 main models: 6,156 responses on T and L (342 per model = 114 conversations $\times$ 3) and 4,266 on O; primary-grader judgments 110,808 on T and L (6,156 per model) and 52,530 on O.
Including the 8 omitted models (26 in total): 8,892 + 6,162 responses and 160,056 + 77,250 judgments.
The secondary grader (Gemini 3.8 Flash low, batch API) made 30,780 judgments (9 main and 6 additional models, one sample each).

\paragraph{Compute.}
Local GPU time (request-processing time on one GPU): 6.9 hours for response generation (longest Weblab-MedLLM, 1.83 h) and 42.9 hours for all primary-grader runs (24.3 h for the runs used in the analysis, 17.4 h for superseded runs and re-grading, 1.2 h for the gold set and meta-evaluation; about 280,000 calls).
Primary-grader throughput was about 7,300 judgments per hour (median over runs; range 3,500--10,200; about 4,000 for models with long responses).

\paragraph{Commercial API cost} at list prices (batch at 50\%; not reconciled with invoices): 16.3 USD for response generation (Opus 9.1, Sonnet 3.3, Sol 1.9, Gemini 1.9, Luna 0.1), 28.7 USD for the secondary grader (56.1M input, 2.7M output, 1.5M reasoning tokens), and 15.0 USD for the grader-candidate comparison; 60.0 USD in total.
Classification calls: 3,440 for the reason-for-change labels used in the paper, 7,441 for the axis re-assignment (Qwen3.8-Flash-Next 17.7 h and DeepSeek 3.0 h on the shared lab API), and 6,002 for exploratory labeling.
Not available: GPU time for DeepSeek V4 Flash and Qwen3.8-Flash-Next (shared lab API), cost of early trial runs without call records, and wall-clock time including server start-up.

\end{document}